\documentclass{article}

\usepackage{natbib}
\usepackage{caption}
\usepackage{subcaption}
\usepackage{hyperref}
\usepackage{authblk}
\usepackage{graphicx}
\usepackage{amsmath, amssymb}
\usepackage{hyperref}

\usepackage{colortbl}
\usepackage[shortlabels]{enumitem}
\usepackage{anyfontsize}
\usepackage{appendix}
\usepackage[a4paper,margin=3cm,marginparwidth=1.75cm]{geometry}
\graphicspath{{images/}, {./}}
\usepackage[table]{xcolor}
\usepackage{xr}

\title{DIRESA, a distance-preserving nonlinear dimension reduction technique based on regularized autoencoders}
\author[1]{Geert De Paepe \thanks{geert.de.paepe@vub.be}}
\author[1,2]{Lesley De Cruz\thanks{lesley.de.cruz@vub.be}}

\affil[1]{Department of Electronics and Informatics, Vrije Universiteit Brussel, Brussels, Belgium}
\affil[2]{Observations Scientific Service, Royal Meteorological Institute, Brussels, Belgium}
\date{\today}

\begin{document}
\maketitle

\abstract{
In meteorology, finding similar weather patterns or analogs in historical datasets can be useful for data assimilation, forecasting, and postprocessing.  In climate science, analogs in historical and climate projection data are used for attribution and impact studies. However, most of the time, those large weather and climate datasets are nearline. This means that they must be downloaded, which takes a lot of bandwidth and disk space, before the computationally expensive search can be executed. We propose a dimension reduction technique based on autoencoder (AE) neural networks to compress the datasets and perform the search in an interpretable, compressed latent space.
A distance-regularized Siamese twin autoencoder (DIRESA) architecture is designed to preserve distance in latent space while capturing the nonlinearities in the datasets.
Using conceptual climate models of different complexities, we show that the latent components thus obtained provide physical insight into the dominant modes of variability in the system.
Compressing datasets with DIRESA reduces the online storage and keeps the latent components uncorrelated, while the distance (ordering) preservation and reconstruction fidelity robustly outperform Principal Component Analysis (PCA) and other dimension reduction techniques such as UMAP or variational autoencoders.}

\begin{center}
\textbf{Significance Statement}
\end{center} This paper introduces a novel, deep learning based dimension reduction method called DIRESA. We believe that this method has the potential to complement the widely used Principal Component Analysis. It gives the user insight into the dominant modes of variability in the system while offering excellent reconstruction fidelity and distance preservation in latent space. 

To increase the impact of this method and facilitate its adoption by the scientific community, we have made it available in a user-friendly Python package. The DIRESA method could thus become a standard tool for many atmospheric and climate scientists.

\section{Introduction}

Today, weather and climate models and observation systems generate unprecedented volumes of data. For example, the Coupled Model Intercomparison Project phase 6 (CMIP6) represents more than 20 PB of climate model output data \citep{li2023big}, and the world's Earth Observation data collections grow by around 100 PB per year \citep{wilkinson2024}.

These unwieldy archives of Earth system data provide enormous opportunities as well as challenges. They are usually stored in formats such as NetCDF, which has natively supported zlib-based compression since version 4.0. This compression technique does not consider the specific context of climate data, and as a result, many more bits of information are stored than are meaningful or necessary \citep{klower2021compressing}. Classical image or video compression techniques may retain some interpretability but are optimized for specific targets, such as the human visual system. The compressed representation learned by neural image compression methods is more adaptive but generally is not interpretable, nor does it allow for similarity search in the compressed domain \citep{mishra2022image}. Compression can also be done by overfitting a coordinate-based neural network and taking the resulting parameters as a compact representation of the original grid-based data \citep{huang2022compressing}, but such an approach still lacks interpretability and distance preservation.

To allow for the full potential of this vast treasure of data to be unlocked, an approach is needed to reduce these datasets in a domain-aware and physically meaningful way. The approach presented in this work uses a Siamese twin AE with regularization terms to anchor the latent space to meaningful properties such as distance (ordering) preservation and statistical independence. The first property is crucial for using algorithms based on distances, such as $k$-means and analogs, in the reduced latent space. The second property reduces redundancies between components and promotes orthogonality in the latent space, ensuring that the latent components represent distinct and independent features with a higher potential for interpretability.
 
A major application that benefits from long records of reanalysis and climate model data is the search for analogs. Methods based on analogs, which rely on the retrieval of similar weather patterns, are used for forecasting \citep{vandendool1989new}, as Lorenz already proposed in 1969 \citep{lorenz1969atmospheric}, or for the calculation of local attractor properties, such as the instantaneous dimension, a proxy for predictability \citep{faranda2017dynamical}. In analog data assimilation, the analogs can be searched in historical data archives \citep{lguensat2017analog} or constructed via generative machine learning methods, such as variational AE \citep{yang2021machine} or generative adversarial networks. Besides statistical and dynamical downscaling, higher resolution weather and climate information can be obtained using analog methods \citep{zorita1999analog,ghilain2022large} or neural network-based techniques \citep{rozoff2022comparison}. Finally, analogs are a cornerstone of a novel climate change attribution approach that accounts for changes in atmospheric circulation patterns \citep{faranda2022climate}.

As a downside to the growing datasets, however, looking up analogs in these vast records becomes increasingly computationally expensive as the dimensionality of the data and length of the datasets increase. As looking up analogs by minimizing Euclidian distance can be practically infeasible, many approaches have been proposed to quickly retrieve similar patterns of one or more atmospheric variables. These are based on dimensionality reduction (DR) techniques such as the popular classical PCA method. To capture the nonlinearity of the data, other techniques were developed \citep{van2009dimensionality}. UMAP (Uniform Manifold Approximation and Projection) is one such technique \citep{mcinnes2018umap}, which can be combined with advanced search algorithms to find the analog in a fast and accurate way \citep{franch2019mass}. DR is also combined with other data analysis techniques, such as clustering \citep{neal2016flexible, chattopadhyay2020predicting}, or topological techniques, such as persistent homology \citep{strommen2022topological}, for finding weather regimes. 

Autoencoders are another technique, based on deep learning, to produce a lower-dimensional latent representation for a dataset or distribution \citep{manning2018pca}. However, they lack the statistical independence of the latent space components and the ordering of these components. The latter can be overcome by adding a MaskLayer \citep{royen2021masklayer}. The correlation between the components is minimized by adding a covariance regularization term on the latent space layer \citep{ladjal2019pca}. 

\begin{figure}[ht]
\centering
\includegraphics[scale=0.3]{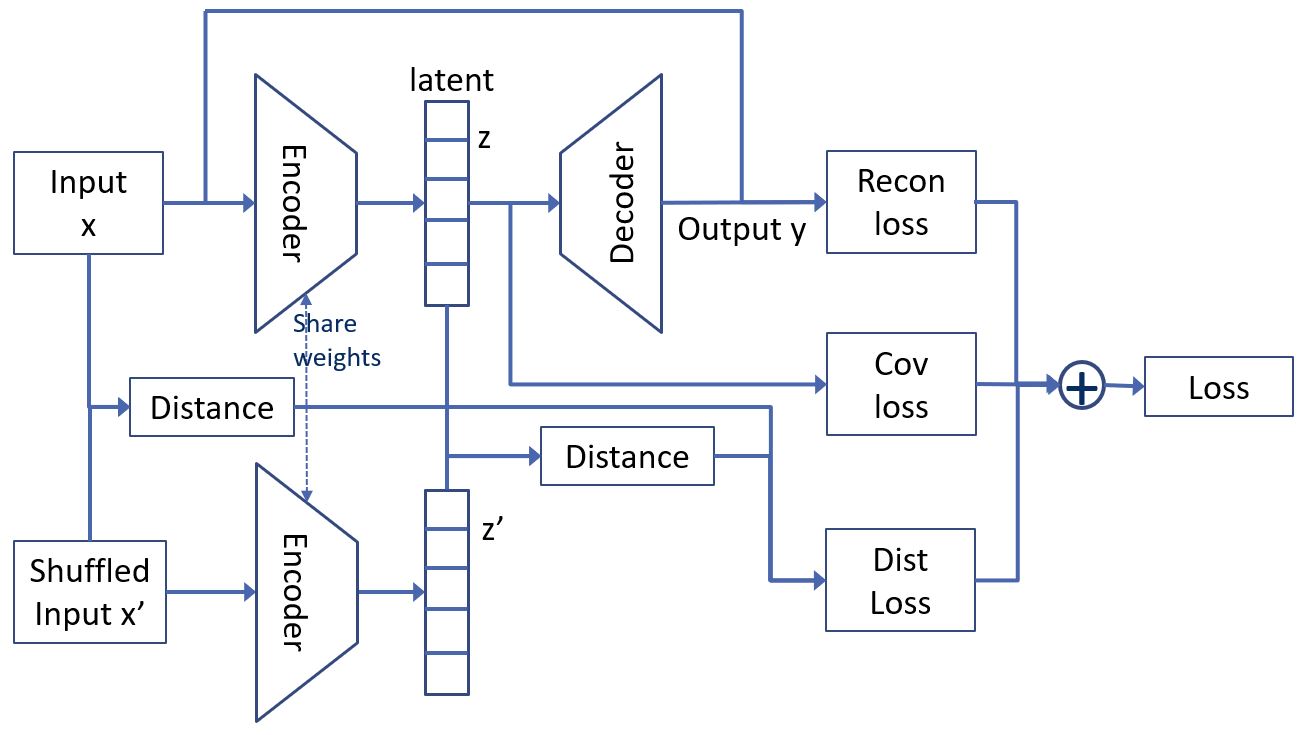}
\caption{DIRESA: Distance regularized siamese twin AE}
\label{fig:siamtwin}
\end{figure}

When using autoencoders, the Euclidian distance between two samples in the original space is not necessarily correlated with the distance in latent space. Our architecture uses a Siamese twin AE (see Figure \ref{fig:siamtwin}) to preserve distance (ordering) in latent space. The two encoder branches share the weights and produce two different latent space representations. These are used for the distance loss term, similar to the invariance term used by \cite{bardes2021vicreg}, which ensures that the distance between samples in the dataset is reflected (preserved or correlated) in the distance between latent representations of those samples. The Siamese twin architecture is also used in AtmoDist for the task of predicting temporal distance between samples to obtain an informative latent representation of the atmospheric state \citep{Hoffmann_Lessig_2023}.
    
To better understand how the latent space relates to the dynamic variables, we demonstrate our method for a hierarchy of conceptual models, namely the well-known Lorenz '63 model and the MAOOAM coupled ocean-atmosphere model, before applying it to reanalysis and state-of-the-art climate model data. Starting with low-dimensional, well-understood models allows us to visualize the latent space and investigate the impact of the various terms of the loss function.

\section{Data}
\subsection{Lorenz '63}

The Lorenz '63 system  \citep{lorenz1963deterministic} is a 3-dimensional simplified model for atmospheric convection. This system of three ordinary differential equations (ODE) displays chaos and features a strange attractor with the famous butterfly-wings shape for specific ranges of its parameters \citep{palmer1993extended}. The 3-dimensional Lorenz butterfly is compressed into a 2-dimensional latent space by the different DR techniques. The system of ODEs with parameters $\sigma=10$, $r=28$, and $b=8/3$ was integrated using the 4th-order Runge-Kutta numerical scheme, from starting point (1,0,1) with a timestep of 0.0025. The first 1000 points are ignored as they are considered part of the transient. The next 100,000 steps are split into a training dataset (first 80,000), a validation (next 10,000), and a test dataset (last 10,000). Before feeding the data into the DR methods, it is scaled to $[0,1]$. Important to remark is that the Hausdorff fractal dimension of the Lorenz butterfly is about 2.06, so it is theoretically impossible to map the 3-dimensional dataset into two dimensions while retaining all the information.

\subsection{Modular Arbitrary-Order Ocean-Atmosphere Model}

MAOOAM (Modular Arbitrary-Order Ocean-Atmosphere Model) is a quasigeostrophic coupled ocean-atmosphere model for midlatitudes \citep{de2016modular, VDDG2015}. It features a 2-layer atmosphere that is mechanically and thermodynamically coupled to a shallow-water ocean layer in the $\beta$-plane (linearized Coriolis) approximation. 
The DDV2016 setup was used, with an atmospheric and oceanic resolution of 2x–2y and 2x–4x, resulting in a 36-dimensional model, with ten wavenumbers for the atmospheric stream function and temperature and eight for the ocean stream function and temperature.
The dataset was obtained by integrating the model with the 2nd-order Runge-Kutta numerical scheme (with nondimensional timestep 0.01) for $4 \times 10^8$ timesteps, with a write-out every 100\textsuperscript{th} timestep (1 timestep is $1 / f_0 $ s, with $f_0$ the Coriolis parameter at 45°N, which is about 2.7 hours). The first 1 million points ($10^8$ timesteps) are ignored as part of the transient, required due to the slow oceanic dynamics. For the following 3 million points, which represent 92,000 years of data, the wavenumbers are converted to a grid of 8 x 6 and four channels (for the four variables) and then split up into a training (first 2.4 million), a validation (next 0.3 million), and a test dataset (last 0.3 million).

\section{Methods}
\subsection{PCA and KPCA}

PCA linearly transforms the data into new coordinates, the principal components \citep{pearson1901liii, hotelling1933analysis}. These components are the eigenvectors of the covariance matrix of the dataset. The amount of dataset variation in the direction of each eigenvector determines the ordering of the coordinates. As the covariance matrix is symmetric, the eigenvalues are real, and the eigenvectors are orthogonal. For dimensional reduction, the coefficients of the first principal components can be used as latent representation, as they explain most of the variance in the dataset.

In KPCA, the traditional PCA is combined with kernel methods \citep{scholkopf1997kernel}. The original data points are mapped non-linearly on a high dimensional Hilbert space, where the PCA is applied. The so-called kernel trick is used to avoid working in this higher dimension (and so being highly computationally expensive). The calculations are done in the original space, replacing the standard inner product with a kernel function representing the inner product in the Hilbert space.

\subsection{UMAP}

UMAP is based on manifold learning and topological data analysis \citep{mcinnes2018umap}. It creates a fuzzy topological representation of the data, a weighted graph, by connecting the points within a certain radius, which is calculated locally based on the distance to the nearest neighbors. A stochastic gradient descent algorithm with a cross-entropy loss function searches for a low-dimensional representation with the closest possible similar fuzzy topological structure. 

An important hyperparameter is the number of neighbors used to construct the initial graph. Low values will focus more on the local details, while high values will better represent the big picture. A second hyperparameter is the minimum distance between points in the latent space. A low value will pack similar points closer together, while with a large value, similar points will be further apart and focus more on the global picture.

\subsection{Autoencoders}

An AE is a deep-learning artificial neural network (ANN) that consists of two parts: an encoder that compresses the data into a latent representation and a decoder that decompresses this representation into its original form \citep{kramer1991nonlinear}. The AE is trained by minimizing the reconstruction loss, which ensures that the output of the decoder is as similar as possible to the encoder's input. We use the MSE between the encoder's input and the decoder's output as reconstruction loss. After the training and testing of the neural network are done, the original dataset can be compressed with the encoder.

The power of AEs comes from the fact that neural networks use non-linear activation functions, allowing them to capture more complex relationships than PCA, which is a linear transformation. Contrary to PCA, however, the latent representation usually lacks interpretability, and its components are not independent or ordered in importance.

In a VAE, the latent components are not single numbers but probability distributions, usually Gaussian distributions \citep{kingma2022autoencoding}. The encoder has three output layers: one for the means, one for the variances (or logarithm of the variances) of the distributions, and one random sampling taken from those distributions that feed into the decoder submodel. A VAE is a generative method, as each sample of the latent distributions produces a different output after decoding. A VAE has two loss functions. The first one is the reconstruction loss, as in a standard AE. The second loss function contains a Kullback–Leibler divergence regularization term (KL loss), which forces the latent distributions to be standard normal distributions. The total loss function is the weighted average of both losses.

\begin{figure}[h]
\centering
\includegraphics[width=0.8\textwidth]{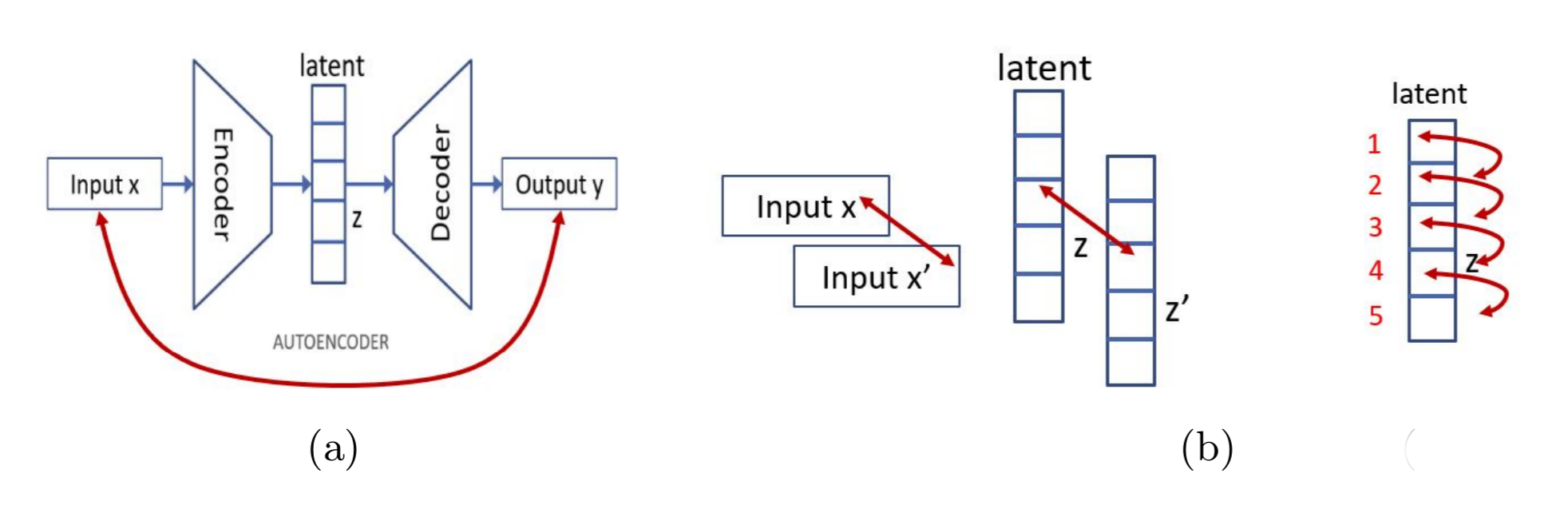}
\caption{(a) Requirement for the AE: Reconstructed data as close as possible to the original. (b) DR requirements: Distance in original space is related to distance in latent space, and latent components are independent and ordered in terms of importance.}
\label{fig:AErequirements}
\end{figure}

\subsection{DIRESA} 

Finally, we present the newly developed DIRESA DR technique. As one of the goals of this technique is to search for similar patterns, the distance between points in the original space should be reflected in their latent-space counterparts. Moreover, to provide insight into a system's dynamics, the latent space should relate to a hierarchy of modes of variability, a kind of non-linear generalization of PCA.
The AE should fulfil the following requirements (Figure \ref{fig:AErequirements}):

    \begin{enumerate} 
        \item Reconstructed dataset as close as possible to the original dataset;
        \item The distance between points in the original space is reflected in their distance in latent space;
        \item The latent components are independent of each other;
        \item The latent components are ordered in terms of importance.
    \end{enumerate}

The last three requirements focus on capturing the essential structure and relationships within the data in a lower-dimensional space, while the first one is specific to autoencoders. The DIRESA architecture (Figure \ref{fig:siamtwin}) fulfils the above requirements. The original dataset $\mathbf{X}= \left(\mathbf{x}_i\right)_{i=1\ldots n}$, with $n$ the number of data points, is encoded by the encoder $ f_{\theta}$, parametrized by $\theta$, into a latent representation $\mathbf{Z}$
\[
f_{\theta}(\mathbf{X}) = \mathbf{Z} \in \mathbb{R}^{n \times L}
\]
where $L$ is the number of latent components. The decoder $ g_{\phi}$, parametrized by $\phi$, decodes the latent dataset back into the original space
\[
g_{\phi}(\mathbf{Z}) = \mathbf{Y}.
\]
As in a standard AE, the reconstruction loss enforces the first requirement. We use the MSE between the original data, fed into the encoder, and the decoder output as reconstruction loss
\[
\mathcal{L}^{MSE}_{recon} = \frac{1}{n} \times \sum_{i=1}^{n} \left\| \mathbf{x}_i -  \mathbf{y}_i \right\|^2.
\]
The reconstruction loss is related to the fraction of variance unexplained $ FVU = 1 - R^2$, with $R^2$ being the coefficient of determination
\[
R^2 = 1 - \frac{\sum_{i=1}^{n} \left\| \mathbf{x}_i -  \mathbf{y}_i  \right\|^2}{\sum_{i=1}^{n} \left\| \mathbf{x}_i -  \mathbf{\bar{x}}  \right\|^2}.
\]
In the case of PCA, $R^2$ corresponds to the percentage of explained variance and the FVU is equal to the unexplained variance.

The twin encoder shares the weights and biases with the original encoder but receives other samples from the data set $\mathbf{X}^{\prime} = \left(\mathbf{x}^{\prime}_j\right)_{j=1,\ldots,n}$ , which are obtained by shuffling the order of the data. The twin encodes a data point 
 $\mathbf{x}^{\prime}$ into $\mathbf{z}^{\prime}$. The random ordering of the shuffled dataset can be done beforehand on the complete dataset (and so does not change during training) or can be executed per batch during training by the data generator. We used the first option here to be independent of the batch size. A distance layer calculates the Euclidian distance $\left\| \mathbf{x} -  \mathbf{x'}  \right\|$ between the original and shuffled input and between the latent components of the two twins $\left\| \mathbf{z} -  \mathbf{z'}  \right\|$. Minimizing the distance loss forces the relationship between those two. For DIRESA, several distance loss functions have been implemented. $\mathcal{L}^{MSE}_{dist}$ uses the MSE 
\[
\mathcal{L}^{MSE}_{dist} = \frac{1}{n} \times \sum_{i=1}^{n}  ( \left\| \mathbf{x}_i -  \mathbf{x'}_i \right\| - \left\| \mathbf{z}_i -  \mathbf{z'}_i  \right\| )^2
\]
meaning that distance is preserved between the original and latent space. The MSLE, which calculates the MSE between $log(\left\| \mathbf{x} -  \mathbf{x'}  \right\|+1)$ and $log(\left\| \mathbf{z} -  \mathbf{z'}  \right\|+1)$, is used for $\mathcal{L}^{MSLE}_{dist}$, focusing more on short than long distances. $\mathcal{L}^{Corr}_{dist}$ uses a distance Pearson correlation loss
\[
\mathcal{L}^{Corr}_{dist} = 1 - Corr(\left\| \mathbf{X} -  \mathbf{X'}  \right\|, \left\| \mathbf{Z} -  \mathbf{Z'}  \right\|)
\]
forcing the correlation between original and latent space distances. The distance Pearson correlation loss is calculated per batch (one figure per batch and not one figure per sample as in an MSE loss function). Therefore, the batch size must be large enough to provide a reliable estimate of the correlation. A last distance loss function is $\mathcal{L}^{LogCorr}_{dist}$, where the correlation loss is computed on the logarithm of the distances plus one.

The independence of the latent components is forced by the covariance loss
\[
\mathcal{L}_{cov} = \frac{1}{L \times (L-1)} \times \sum_{i \neq j} cov^2_{i,j}(\mathbf{Z}) 
\]
which is the normalized squared sum of the non-diagonal terms of the covariance matrix. The denominator keeps the loss function independent of the latent space dimension. The covariance matrix is calculated by batch, so again, it is essential to have the batch size big enough so that the batch covariance is a good approximation of the dataset covariance. 

The total loss is a weighted average of the three different loss components. For lowering the weight factor tuning effort, annealing \citep{behrens2022non} is used for the covariance loss, meaning that the weight factor is 0 at the start and gradually increases. Annealing stops when the covariance loss reaches a point, where we consider the latent components independent. The covariance regularized AE (CRAE) \citep{ladjal2019pca} is similar to DIRESA but with only the covariance loss and no distance loss (and so no twin encoder).

The last requirement, the ordering of the latent components, could be imposed by a \textit{Masklayer}, but in practice, it was found to hamper the training of the neural network. Therefore, the ordering is not imposed during training but is calculated afterwards, based on the $R^2$ score. The procedure for this is described in Appendix E.
  
In Appendix E, an ablation study demonstrates the usefulness of the different loss functions. The ablation study shows that the reconstruction loss can be left out, which yields a simplified DIRESA architecture. However, without the reconstruction loss, there is no decoder for the latent space dataset, and without a decoder, the ordering of the components, according to the procedure in Appendix D, cannot be calculated.

\begin{table}
\centering
\begin{tabular}{||l | l | l | l | l ||} 
 \hline
 DR	       & Recon MSE         & Cov Loss       &   Dist Loss      &   KL Loss \\
 \hline\hline
 PCA		       &   0.00191           &                 &                   & \\
 KPCA           &   0.000269          &                 &                   & \\
 UMAP	       &   0.000282          &                 &                   & \\
 AE             &   1.16e-05          &                 &                   & \\
 BNAE           &   5.21e-05          &                 &                   & \\
 CRAE           &   1.91e-05          &  7.95e-08       &                   & \\
 VAE            &   8.82e-05          &                 &                   & 5.20e-06 \\
 DIRESA\textsubscript{MSE}        &   0.000114          &  1.31e-05       &  0.000219         & \\
 DIRESA\textsubscript{Corr}       &   0.000125          &  3.24e-06       &  0.00125          & \\
 \hline
\end{tabular}
\caption{Model losses the Lorenz '63 test dataset}
\label{table:LossLorenz}
\end{table}

\begin{figure}[ht]
\centering
\includegraphics[width=\textwidth]{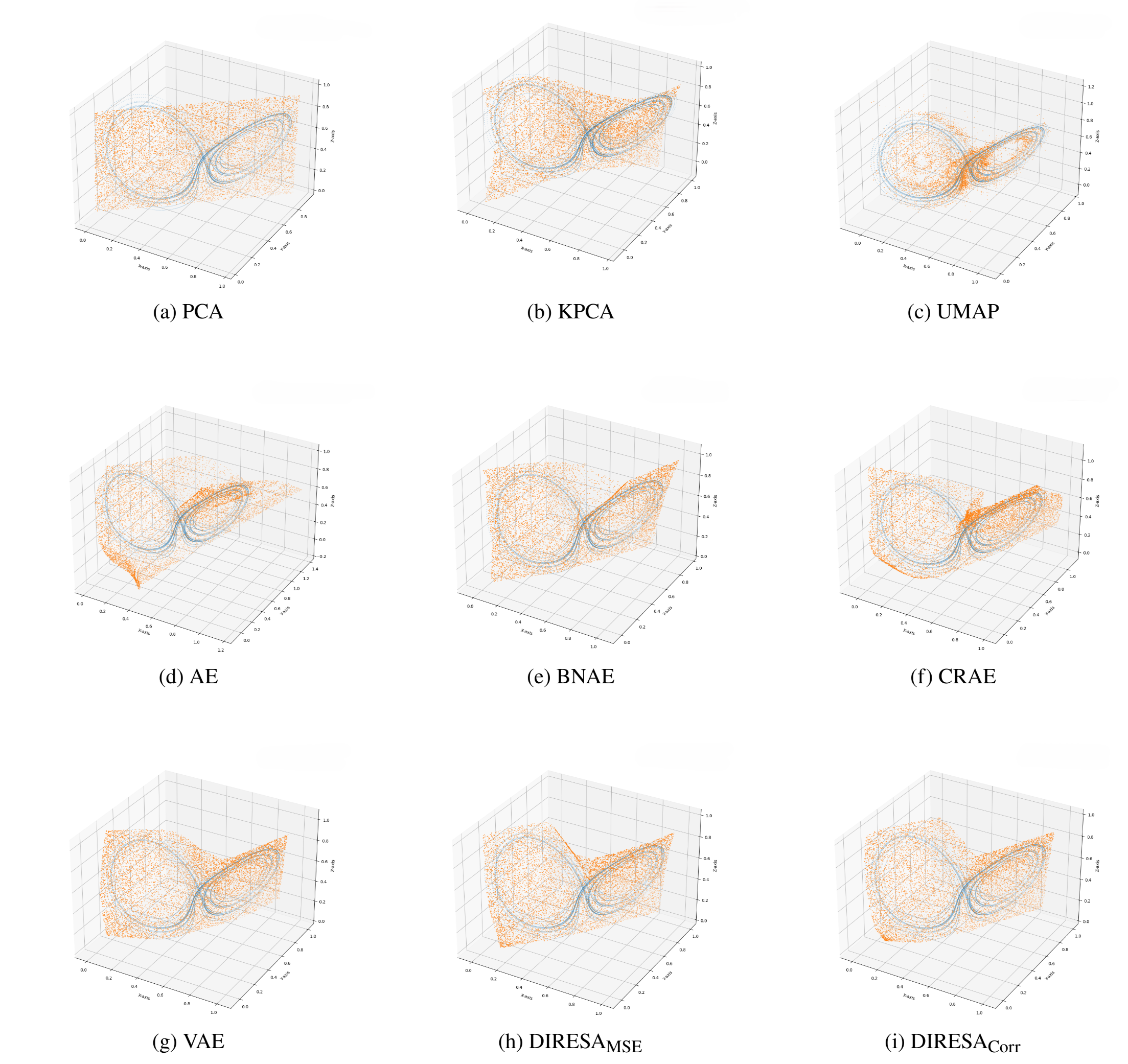}
\caption{Lorenz test dataset (blue) next to the decoded random uniform latent components (orange) for the different DR techniques}
\label{fig:declatentlorenz}
\end{figure}

\begin{figure}[ht]
\centering
\includegraphics[width=\textwidth]{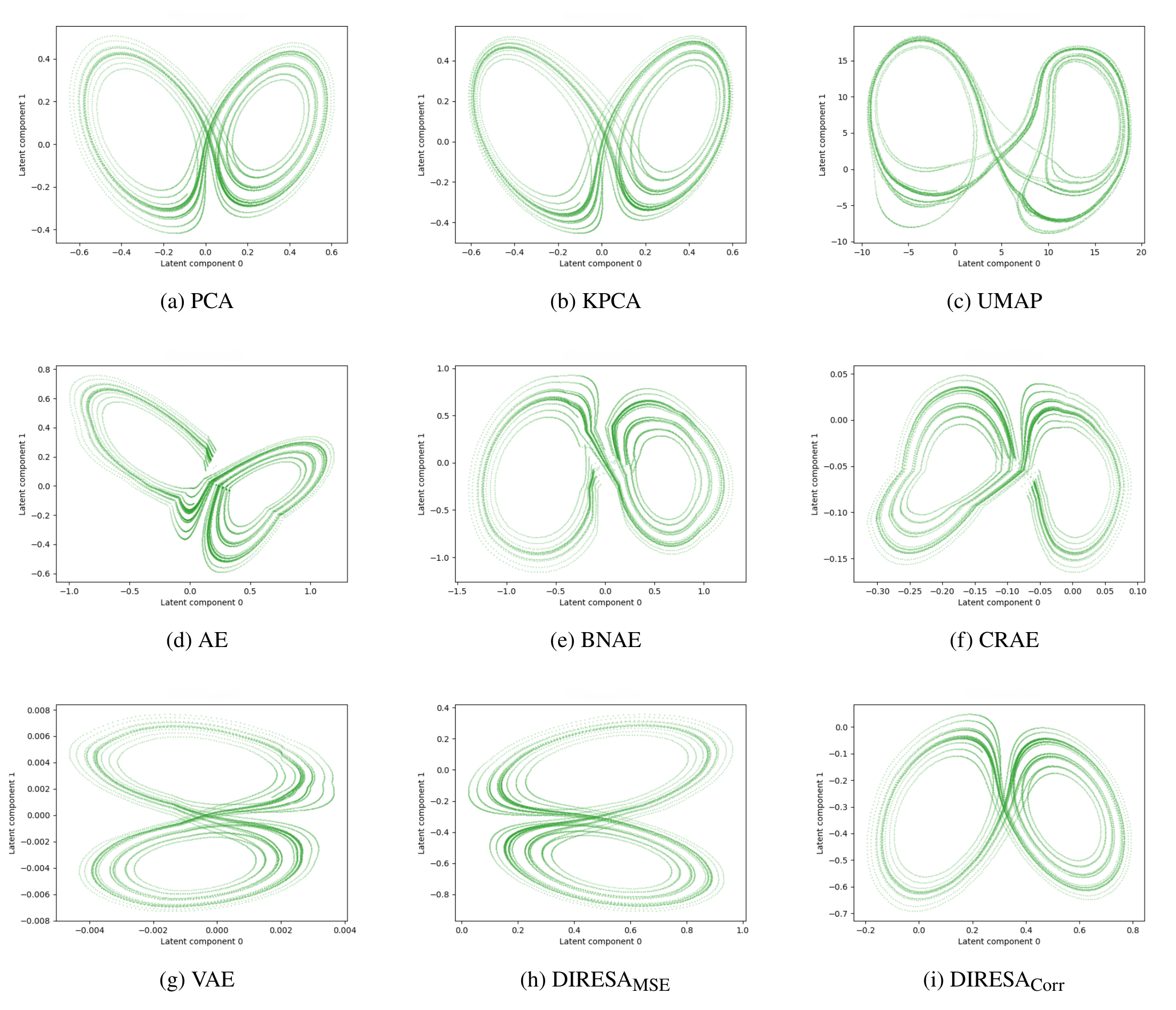}
\caption{Encoded Lorenz test dataset for the different DR techniques}
\label{fig:latentlorenz}
\end{figure}
    
\section{Results}
\subsection{Lorenz '63}

In our benchmark, we use nine different DR techniques to compress the 3-dimensional Lorenz butterfly into a 2-dimensional latent space: Principal Component Analysis (PCA), kernel PCA (KPCA), Uniform Manifold Approximation and Projection (UMAP), and six ANNs: AE, AE with batch normalization on the latent space (BNAE), AE with covariance regularization on the latent space (CRAE), variational AE (VAE) and two distance regularized siamese twin AE (DIRESA). 

The hyperparameters for the nine different DR techniques are listed in Table \ref{table:ModelConfLorenz}. Hyperparameter tuning is done using simulated annealing \citep{kirkpatrick1983optimization} for KernelPCA and UMAP. More information on the hyperparameter tuning can be found in Appendix B. All ANN methods have an encoder submodel with a 3-dimensional input layer, two hidden layers with 40 and 20 fully connected nodes, and an output layer with two fully connected nodes, see Table \ref{table:ModelDiresaLorenz}. This results in an AE and DIRESA ANN with 2045 trainable parameters.

\subsubsection{Evaluation of latent and decoded samples}

The different DR methods are evaluated quantitatively based on the loss function output of the test dataset (see Table \ref{table:LossLorenz}). The properties of the latent space are assessed qualitatively by visualizing the decoding uniform random points in latent space (see Figure \ref{fig:declatentlorenz}) and by visualizing the encoded test dataset in latent space (see Figure \ref{fig:latentlorenz}). In terms of reconstruction loss, PCA performs the worst (because of the linearity), and the latent space is decoded into a flat plane (see Figure \ref{fig:declatentlorenz}a). A standard AE has the lowest reconstruction loss (and so the highest $R^2$ score). Still, the decoded random sample shown in Figure \ref{fig:declatentlorenz}d is irregular and fails to generalize well outside the wings. BNAE, CRAE, and VAE are also doing well from a reconstruction loss point of view. Both DIRESA methods are doing better for the reconstruction loss than KPCA and UMAP, although a trade-off had to be made between the three different loss functions. The decoded latent space curves for both DIRESA variants follow the wings, and outside the wings, the curve is well generalized (see Figure \ref{fig:declatentlorenz}h and \ref{fig:declatentlorenz}i), which is also the case for BNAE (see Figure \ref{fig:declatentlorenz}e) and VAE (see Figure \ref{fig:declatentlorenz}g). Using an \textit{MSE} or a \textit{Corr} distance loss function makes little difference from a reconstruction point of view. Unlike the other methods, the UMAP decoded sample is more scattered (see Figure \ref{fig:declatentlorenz}c) and does not lie in a 2-dimensional manifold. 

Figure \ref{fig:latentlorenz} shows the mapping of the Lorenz test dataset into latent space. The encoded latent components are uncorrelated for all DR methods except for the standard AE (see Figure \ref{fig:latentlorenz}d). As we can see in Figure \ref{fig:latentlorenz}a and \ref{fig:latentlorenz}b, PCA and KPCA order the latent components by variance, which is not the case for the different ANNs, see Figure \ref{fig:latentlorenz}g for the VAE and Figure \ref{fig:latentlorenz}h for DIRESA, where the wing is rotated. The wing structure is well preserved with PCA, KPCA, VAE, and DIRESA methods, which is less the case for UMAP (see Figure \ref{fig:latentlorenz}c) and the AE. With a standard AE, no regularization is done on latent space, so we get different irregular pictures for other training runs. For BNAE, both latent components have a standard deviation of 1, so the wing structure is resized in one direction (see Figure \ref{fig:latentlorenz}e).  CRAE has no distance regularization, so one wing is more prominent here than the other (see Figure \ref{fig:latentlorenz}f).

\subsubsection{Distance ordering preservation} \label{dist}

\begin{table}
\centering
\begin{tabular}{|l || l | l || l | l | l | l |} 
\hline
DR & Corr & LogCorr & Can50 & Pear50 & Spear50 & Ken50 \\
\hline\hline

PCA & \textbf{0.997\textsuperscript{*}}& \textbf{0.996} & \textbf{0.117} & 0.839 & \textbf{0.889} & \textbf{0.834} \\
\hline
KernelPCA & 0.988 & 0.989 & 0.201 & 0.775 & 0.823 & 0.709 \\
\hline
UMAP & 0.918 & 0.892 & 0.300 & 0.711 & 0.751 & 0.603 \\
\hline

AE & 0.924 & 0.936 & 0.257 & 0.805 & 0.800 & 0.649 \\
\hline
BNAE & 0.987 & 0.987 & 0.155 & \textbf{0.853} & 0.874 & 0.769 \\
\hline
CRAE & 0.945 & 0.946 & 0.178 & \textbf{0.859\textsuperscript{*}} & 0.866 & 0.741 \\
\hline

VAE & 0.979 & 0.973 & 0.151 & \textbf{0.858} & 0.874 & 0.768 \\
\hline
DIRESA\textsubscript{MSE} & \textbf{0.997\textsuperscript{*}} & \textbf{0.997\textsuperscript{*}} & \textbf{0.107\textsuperscript{*}} & 0.849 & \textbf{0.896\textsuperscript{*}} & \textbf{0.848\textsuperscript{*}} \\
\hline
DIRESA\textsubscript{Corr} & \textbf{0.997\textsuperscript{*}} & \textbf{0.997\textsuperscript{*}} & \textbf{0.112} & 0.845 & \textbf{0.894} & \textbf{0.843} \\
\hline

\end{tabular}
\caption{Mean distance ordering preservation between original and latent space for the Lorenz '63 test dataset. The \textit{Corr} and \textit{LogCorr} columns show the Pearson correlation for Euclidean and logarithmic distances. The \textit{Can50} column gives the Canberra stability indication (smaller is better), the \textit{Pear50}, \textit{Spear50} and \textit{Ken50} show the Pearson, Spearman, and Kendall correlation with a location parameter of 50. The 3 best values are highlighted in bold; the best one is marked with an asterisk.}
\label{table:LorenzDistPreservationMean50}
\end{table}

\begin{figure}[ht]
\centering
\includegraphics[width=\textwidth]{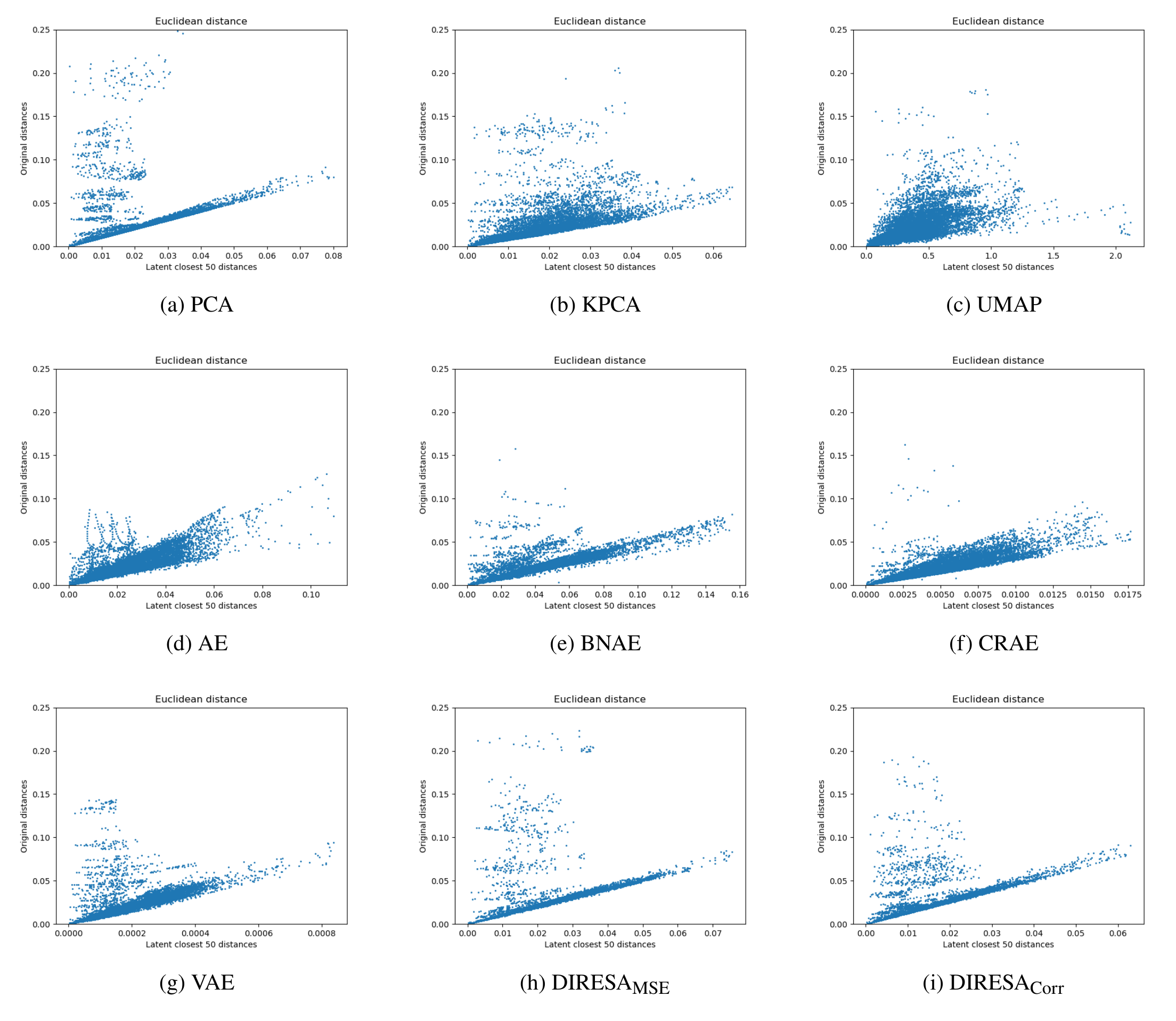}
\caption{Scatter plots of 50 closest distances for 200 random samples of the Lorenz test dataset for for the different DR techniques} 
\label{fig:LorenzCorr50}
\end{figure}

Table \ref{table:LorenzDistPreservationMean50} shows how distance ordering is preserved between original and latent space for the different DR methods. We calculated the distances from all points of the test dataset to all other points and did the same for the corresponding points in latent space. The \textit{Corr} column shows the average Pearson correlation between the Euclidean distances in original and latent space. The \textit{LogCorr} shows the same but for the logarithm of the distances. When searching for analogs, we are only interested in the closest distances. In the right four columns of Table \ref{table:LorenzDistPreservationMean50}, a location parameter 50 has been set, meaning that only the 50 closest distances (in latent space) are considered. The \textit{Can50} shows the mean Canberra stability indicator between the Euclidean distances. The Canberra stability indicator measures the difference between 2 ranked lists (smaller is better). A Canberra stability indicator of about 0 means the lists are ranked very similarly, while two random lists will result in a value of around 1.42. Rank differences at the top of the list are more penalized than differences in the bottom part. The \textit{Pear50}, \textit{Spear50} and \textit{Ken50} columns are showing Pearson's r, Spearman's $\rho$ and Kendall's $\tau$ correlation with a location parameter of 50 ( Appendix Table \ref{table:LorenzDistPreservationMean100} shows the same indices for a location parameter of 100). Table \ref{table:LorenzDistPreservationMedian50} shows the median of the same indicators. The three best scores are shown in red, and the best score is in dark red. The Appendix Table \ref{table:LorenzDistPreservationStdErr50} shows the standard error of the mean for the different indicators. Table \ref{table:LorenzDistPreservationPvalue} shows the 2 sample t-test p-value for the mean distance between DIRESA and PCA. Figure \ref{fig:LorenzCorr50} shows the scatter plots of the 50 closest distances (in latent space) for the different methods for 200 random samples of the test dataset. The points in the two wings' cross-section are where the distance ordering preservation is worst when mapped on a 2-dimensional curve.

\begin{table}
\centering
\begin{tabular}{|l || l | l || l | l | l | l |} 
\hline
DR & Corr & LogCorr & Can50 & Pear50 & Spear50 & Ken50 \\
\hline\hline

PCA & \textbf{0.999\textsuperscript{*}} & \textbf{0.999\textsuperscript{*}} & \textbf{0.050} & \textbf{0.991} & \textbf{0.986} & \textbf{0.920} \\
\hline
KernelPCA & 0.992 & 0.993 & 0.157 & 0.894 & 0.898 & 0.750 \\
\hline
UMAP & 0.936 & 0.894 & 0.280 & 0.745 & 0.791 & 0.623 \\
\hline

AE & 0.938 & 0.947 & 0.233 & 0.845 & 0.833 & 0.670 \\
\hline
BNAE & 0.994 & 0.992 & 0.097 & 0.972 & 0.953 & 0.837 \\
\hline
CRAE & 0.964 & 0.964 & 0.151 & 0.930 & 0.914 & 0.768 \\
\hline

VAE & 0.982 & 0.977 & 0.107 & 0.964 & 0.948 & 0.827 \\
\hline
DIRESA\textsubscript{MSE} & \textbf{0.999\textsuperscript{*}} & \textbf{0.999\textsuperscript{*}} & \textbf{0.040\textsuperscript{*}} & \textbf{0.995\textsuperscript{*}} & \textbf{0.990\textsuperscript{*}} & \textbf{0.935\textsuperscript{*}} \\
\hline
DIRESA\textsubscript{Corr} & \textbf{0.999\textsuperscript{*}} & \textbf{0.999\textsuperscript{*}} & \textbf{0.045} & \textbf{0.994} & \textbf{0.989} & \textbf{0.928} \\
\hline

\end{tabular}
\caption{Median distance ordering preservation between original and latent space for the Lorenz '63 test dataset. Columns and highlights as in Table \ref{table:LorenzDistPreservationMean50}.}
\label{table:LorenzDistPreservationMedian50}
\end{table}

Although all methods perform better than PCA concerning the reconstruction MSE, PCA outperforms almost all for the distance ordering preservation indicators, except for the DIRESA methods (which can also be seen in the scatter plots). KPCA, although grasping the curvature, does worse than PCA for distance ordering preservation. Standard AEs perform worse than the PCA on all scores because they lack regularization on latent space. Batch normalization on latent space significantly improves all scores. Looking at the median score, the DIRESA\textsubscript{MSE} method scores best on all indicators, closely followed by DIRESA\textsubscript{Corr} and PCA. For the mean scores, DIRESA\textsubscript{MSE} scores best on 5 out of 6 indicators, showing that the best way to preserve distance ordering is to preserve the distance itself.

\subsection{MAOOAM}

\begin{figure}[ht]
\centering
\includegraphics[scale=0.3]{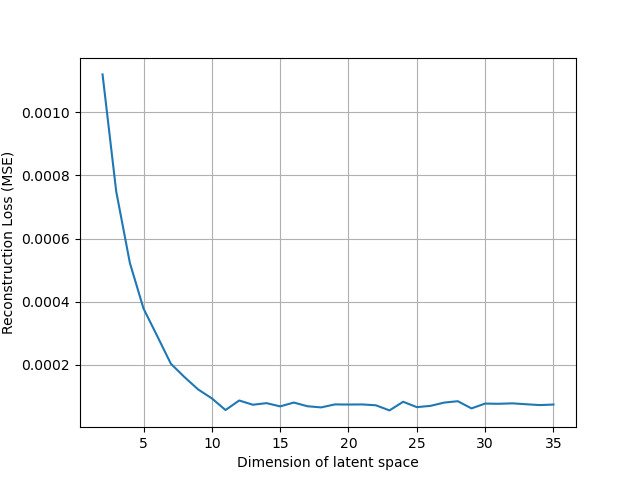}
\caption{MAOOAM: AE Reconstruction loss in function of latent size}
\label{fig:ma_latent_size}
\end{figure}

For MAOOAM, the latent space dimension has been fixed to 10, based on a simulation with a standard AE (see Figure \ref{fig:ma_latent_size}). When lowering the latent space dimension below 10, the reconstruction loss increases fast, which gives us an idea of the dimension of the attractor of the dynamical system.

\begin{figure}[ht]
\centering
\includegraphics[width=\textwidth]{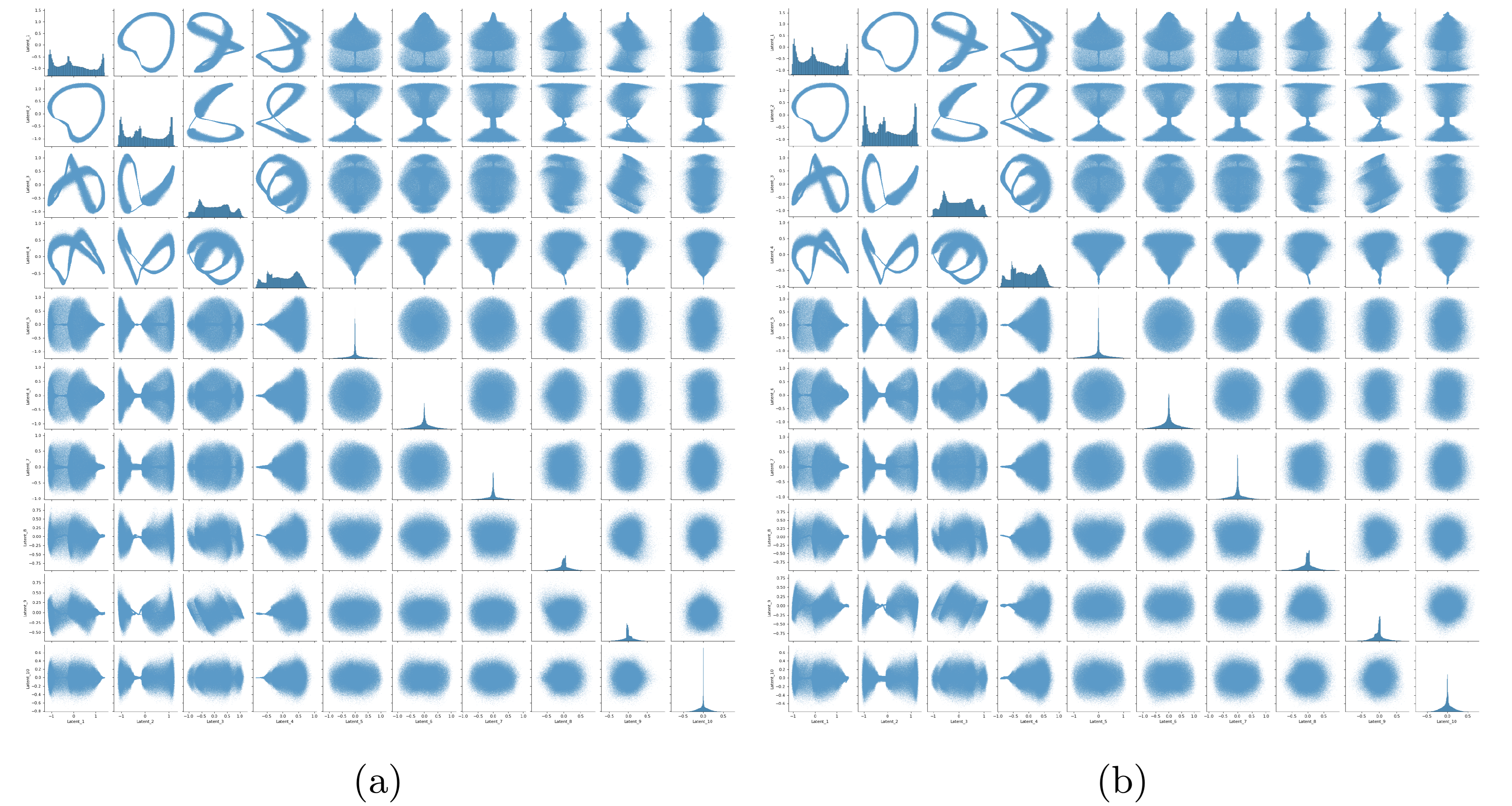}
\caption{Pair plot for MAOOAM latent components for PCA (a)} and DIRESA\textsubscript{MSE} (b)
\label{fig:MA_Pair}
\end{figure}
    
For MAOOAM, we will compare DIRESA with PCA. The complete DIRESA configuration can be found in Appendix B. The encoder of the DIRESA DR methods has one convolutional layer, followed by a MaxPooling and three fully connected layers, see Table \ref{table:ModelDiresaMA}. This results in a DIRESA ANN with 791,314 trainable parameters. The values for all other hyperparameters can be found in \ref{table:ModelConfMA}.

\begin{figure}[ht]
\centering
\includegraphics[width=0.7\textwidth]{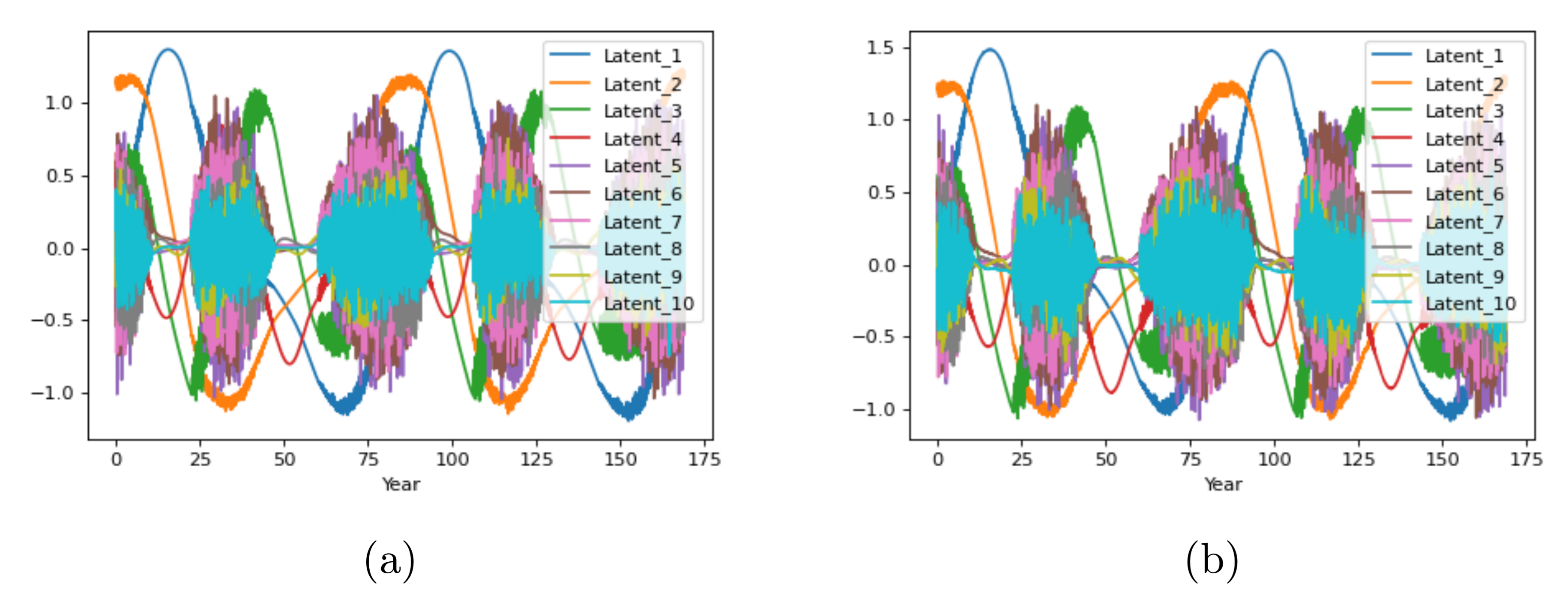}
\caption{Time plot for MAOOAM latent components for PCA (a)} and DIRESA\textsubscript{MSE} (b)
\label{fig:MA_Time}
\end{figure}

\subsubsection{Evaluation of latent and decoded samples}

\begin{figure}[ht]
\centering
\includegraphics[scale=0.42]{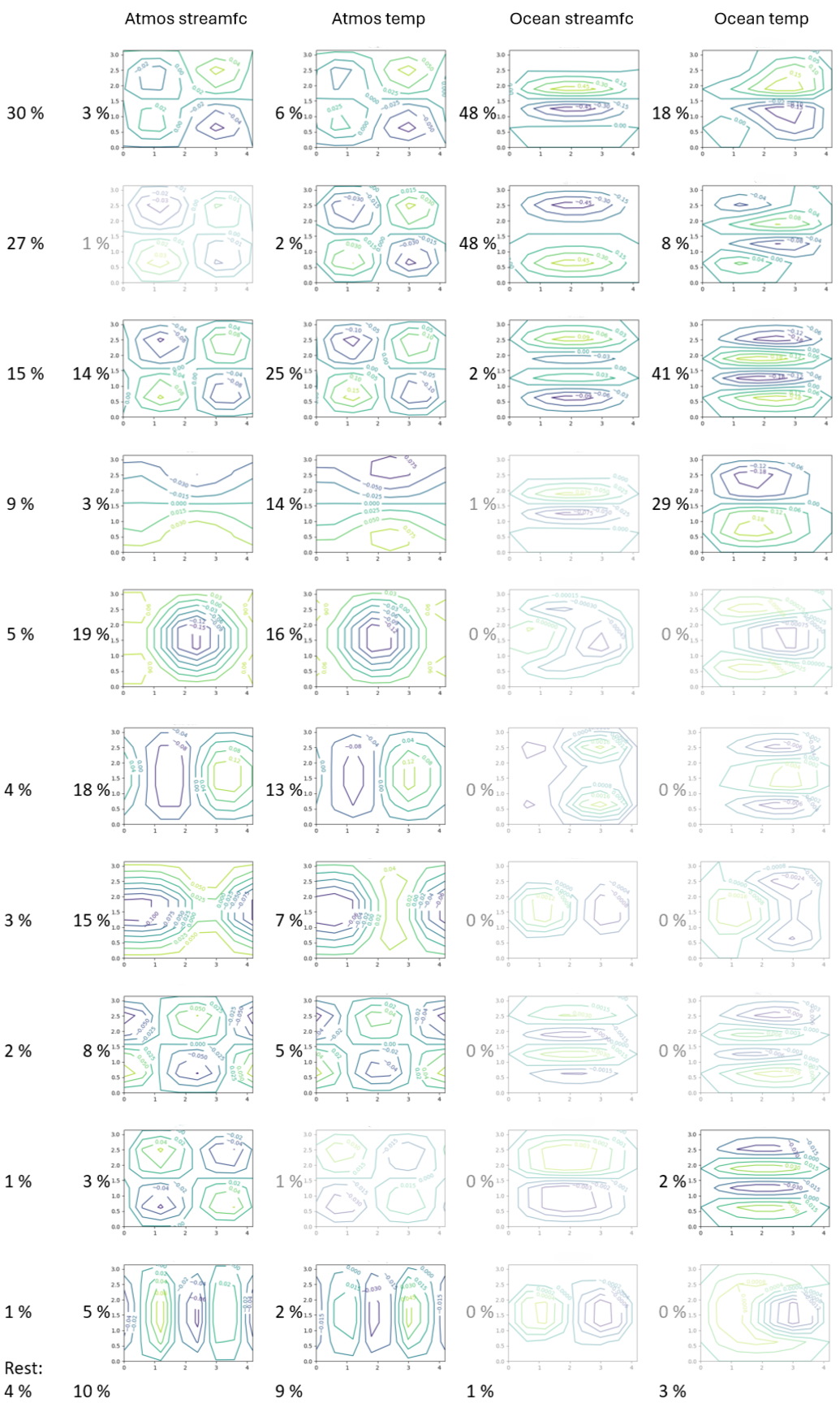}
\caption{MAOOAM: PCA empirical orthogonal functions for the atmospheric stream function, atmospheric temperature, ocean stream function and ocean temperature. Figures show the total explained variance (on the left) and the explained variance per variable. Decoded components for a variable with an explained variance of less than 1.5 are faded.)}
\label{fig:MA_latent_comp_pca}
\end{figure}

\begin{figure}[ht]
\centering
\includegraphics[scale=0.42]{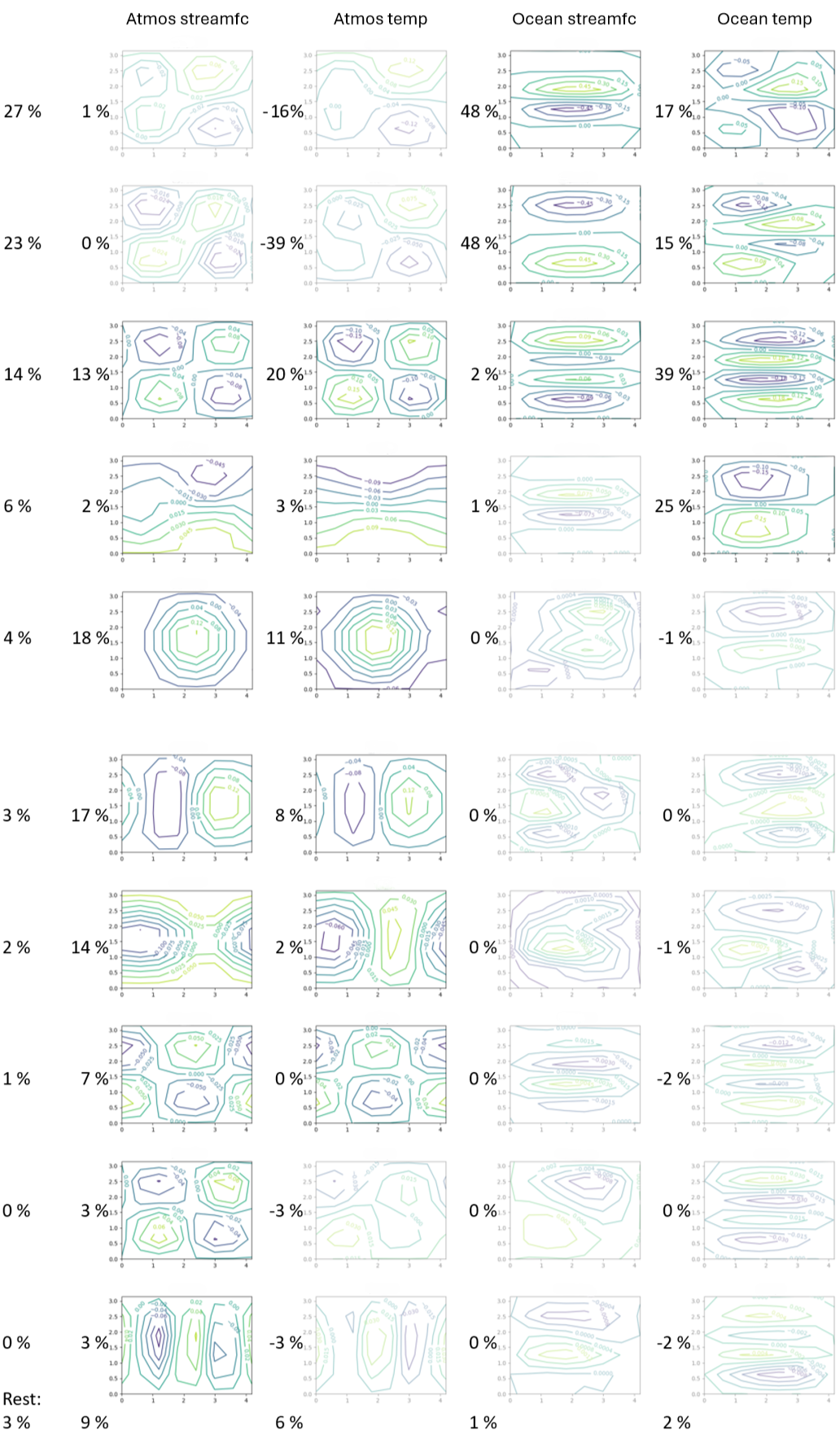}
\caption{MAOOAM: DIRESA decoded latent components for the atmospheric stream function, atmospheric temperature, ocean stream function and ocean temperature. Figures show the total $R^2$ score (on the left) and the $R^2$ score per variable. Decoded components for a variable with an $R^2$ score of less than 1.5 are faded.)}
\label{fig:MA_latent_comp_di}
\end{figure}
    
Figure \ref{fig:MA_Pair} shows the pair plots for the latent components of the test dataset for PCA and DIRESA with MSE distance loss (after sorting). The DIRESA pair plots look very similar to those of PCA. Figure \ref{fig:MA_Time} shows the time plots for the latent components of the test dataset for PCA and DIRESA with MSE distance loss. Again, the DIRESA time plot looks very similar to the PCA's. The first four latent components with the largest $R^2$ score have a low-frequency pattern, while the last six latent components have a high frequency. 
Figure \ref{fig:MA_latent_comp_pca} shows the empirical orthogonal functions (or PCA eigenvectors) for the different variables together with the percentage of explained variance (overall and per variable). The decoded latent components for DIRESA are shown in Figure \ref{fig:MA_latent_comp_di} (the procedure for calculating these is explained in Appendix D). Decoded components for a variable with an $R^2$ score of less than $1.5$ are faded. Again, we notice that the decoded latent DIRESA components (with MSE distance loss) are very similar to the PCA empirical orthogonal functions, except that the $R^2$ score is higher. 
The FVU for PCA is 3.7\%, and for DIRESA 2.8\%, the difference is the biggest for the atmospheric temperature, where the FVU for PCA is 9.3\% and for DIRESA 6.0\%. PCA shows a strong coupling between atmospheric and ocean variables for the first, third and fourth component (with an $R^2$ score of more than 5\% for at least one of the atmospheric and ocean variables). A small coupling is shown for the second components, with an $R^2$ of 2\% for the atmospheric temperature, 48\% for the ocean stream function and 8\% for the ocean temperature.
DIRESA shows only a strong coupling for the third component and a low coupling for the fourth component. The first two components have a cumulative $R^2$ score (see Appendix D) of more than 95\% for the ocean stream function (for both PCA and DIRESA). For the ocean temperature, the first four components are needed for a similar cumulative $R^2$ score. DIRESA shows no coupling between atmospheric and ocean variables for the high-frequency components, similarly to PCA.

\begin{table}
\centering
\begin{tabular}{|l || l | l || l | l | l | l |} 
\hline
DR & Corr & LogCorr & Can50 & Pear50 & Spear50 & Ken50 \\
\hline\hline

PCA & \textbf{0.999\textsuperscript{*}} & 0.998 & 0.570 & 0.432 & 0.398 & 0.292 \\
\hline

DIRESA\textsubscript{MSE} & \textbf{0.999\textsuperscript{*}} & \textbf{0.999\textsuperscript{*}} & 0.565 & 0.429 & 0.401 & 0.292 \\
\hline
DIRESA\textsubscript{MSLE} & \textbf{0.999\textsuperscript{*}} & 0.998 & \textbf{0.518\textsuperscript{*}} & \textbf{0.490\textsuperscript{*}} & \textbf{0.453\textsuperscript{*}} & \textbf{0.338\textsuperscript{*}} \\             
\hline
DIRESA\textsubscript{Corr} & \textbf{0.999\textsuperscript{*}} & \textbf{0.999\textsuperscript{*}} & 0.553 & 0.453 & 0.417 & 0.304 \\
\hline
DIRESA\textsubscript{LogCorr} & \textbf{0.999\textsuperscript{*}} & \textbf{0.999\textsuperscript{*}} & \textbf{0.534} & \textbf{0.472} & \textbf{0.443} & \textbf{0.328} \\
\hline

\end{tabular}
\caption{Mean distance ordering preservation between original and latent space for 200 random samples of the MAOOAM test dataset. Columns and highlights as in Table \ref{table:LorenzDistPreservationMean50}.}
\label{table:MADistPreservation}
\end{table}

\subsubsection{Distance ordering preservation}

Table \ref{table:MADistPreservation} shows the distance (ordering) preservation for PCA and DIRESA with four different distance loss functions: MSE, MSLE, correlation, and correlation of the logarithm of the distances. As it is computationally expensive to calculate distances from all points of the test dataset to all other points, we took 200 random samples from the test dataset and calculated the distance to all other points. 200 random samples are enough to distinguish the best DIRESA methods (in this case, DIRESA\textsubscript{MSLE} and DIRESA\textsubscript{LogCorr}) from PCA with a good p-value (see Appendix Table \ref{table:MADistPreservationPvalue}). The figures are shown for the same KPIs as in paragraph \ref{dist}. Looking at all the distances, near and far, PCA and DIRESA methods are performing equally for the \textit{Corr} KPI, and three out of the four DIRESA methods are giving slightly better figures for the \textit{LogCorr} KPI (which provides more emphasis on the short distances). Looking at the 50 closest distances, all DIRESA methods outperform PCA for \textit{Can50} and \textit{Spear50}, and three out of four DIRESA methods outperform PCA for the \textit{Pear50} and \textit{Ken50} KPIs. Methods with a logarithmic distance loss function yield the best results.

\subsection{Conclusion}

A deep ANN DR method, called DIRESA, has been developed to capture nonlinearities while preserving distance (ordering) and producing statistically independent latent components. The architecture is based on a Siamese twin AE, with three loss functions: reconstruction, covariance, and distance. An annealing method is used to automate the otherwise time-consuming process of tuning the different weights of the loss function. 

DIRESA has been compared with PCA and state-of-the-art DR methods for two conceptual models, Lorenz '63 and MAOOAM, and significantly outperforms them in terms of distance (ordering) preservation KPIs and reconstruction fidelity. We have also shown that the latent components have a physical meaning as the dominant modes of variability in the system. DIRESA correctly identifies the major coupled modes associated with the low-frequency variability of the coupled ocean-atmosphere system. 

In addition to the present study, we provide an open-source Python package, based on Tensorflow \citep{tensorflow2015-whitepaper}, to build DIRESA models with convolutional and/or dense layers with one line of code \citep{depaepe2024diresa}. On top of that, the package allows the use of custom encoder and decoder submodels to build a DIRESA model. The DIRESA package acts as a meta-model, which can use submodels with various kinds of layers, such as attention layers, and more complicated designs, such as graph neural networks or hybrid quantum neural networks. Thanks to its extensible design, the DIRESA framework can handle more complex data types, such as three-dimensional, graph, or unstructured data. 

In the future, we will use the DIRESA DR method on real-world datasets such as weather radar data or reanalysis data. Thanks to the flexibility of the neural encoder, we expect the highest benefit compared to PCA and other traditional DR methods on datasets originating from systems with an attractor spanning a strongly nonlinear, lower-dimensional subspace. Large-scale atmospheric dynamics, quantified by daily sea level pressure fields, are an example of such a system \citep{faranda2017dynamical}. For a purely stochastic system in which the attractor does not span a lower-dimensional subspace, we do not expect DIRESA to perform better than other DR methods. A prerequisite for using DIRESA is that the training dataset is large enough, which is less necessary for PCA.

We believe that DIRESA's flexibility and robust performance give it all the assets to become a useful tool to distil meaningful low-dimensional representations from the ever-increasing volumes of high-resolution climate data.

\section{Data availability}
The data used in this study is available on the open data portal Zenodo \citep{depaepe2024dataset}.

\section{Code availability}
The DIRESA package source code is available on the open data portal Zenodo \citep{depaepe2024diresa}. Other programs and scripts are available from the corresponding author upon reasonable request. Documentation can be consulted at \verb|https://diresa-learn.readthedocs.io|.

\section{Acknowledgements} 
We thank Steven Caluwaerts for the opportunity to start working on this subject as part of the Postgraduate Studies in Weather and Climate Modeling at Ghent University. We would also like to thank Herwig De Smet for reviewing this document and Alex Deckmyn for using his code for solving the Lorenz '63 equations. LDC acknowledges support from the Belgian Science Policy Office (BELSPO) through the FED-tWIN program (Prf-2020-017) and project B2/233/P2/PRECIP-PREDICT. GDP and LDC acknowledge support from VUB through the project SRP74: LSDS. The funders played no role in the study design, data collection, analysis and interpretation of data, or the writing of this manuscript.  

\section{Author contributions}
GDP: Data, Conceptualization, Investigation, Analysis, DIRESA Architecture, Software Development, Methodology, Writing, Visualization, Reviewing, Editing. LDC: Conceptualization, Analysis, Methodology, Writing, Reviewing, Editing, Supervision. 

\section{Competing interests}

All authors declare no financial or non-financial competing interests. 
\clearpage 

\appendix
\appendixpage
\renewcommand{\thesection}{\Alph{section}}

\section{Software and hardware details}\label{sec:pack}

We use the Sklearn-pandas machine learning library version 2.2.0 \citep{scikit-learn} to execute the PCA and KPCA calculations and the umap-learn python package version 0.5.3 \citep{mcinnes2018umap-software} for the UMAP machine learning. We use the TensorFlow/Keras software library version 2.7.1 for the ANN DR models. 

All ANN calculations were performed on NVIDIA A100 GPU nodes and AMD EPYC 7282 (Zen2) CPUs or on Tesla P100 GPU nodes and INTEL E5-2650v4 (Broadwell) of the VUB Hydra high-performance computing infrastructure. The PCA, KernelPCA, and UMAP calculations were performed on INTEL E5-2680v4 (Broadwell) or on INTEL Xeon Gold 6148 (Skylake) of the VUB Hydra high-performance computing infrastructure.

\section{Hyperparameter tuning} \label{sec:hyp}
\subsection*{Lorenz '63 DR hyperparameter tuning}

\begin{table}
\centering
\begin{tabular}{||l | l | l | l ||} 
 \hline
 DR	           &    Parameter Type               &  Initial    &  Step Std Dev \\ 
 \hline\hline
 
 KPCA           &   gamma                         &  1.0        &  0.2 \\
 UMAP           &   min\_dist; n\_neighb	         &  0.1; 150   &  0.02; 30 \\
 \hline\hline
 DR	           &   Annealing Parameter           &  Value      &   \\ 
 \hline\hline
 KPCA, UMAP     &   T\textit{min}                 &  1e-5       &   \\
 KPCA, UMAP     &   T\textit{max}                 &  0.0003     &   \\
 UMAP           &   n\_iterations                 &  200        &  \\
 KPCA           &   n\_iterations                 &  100        &  \\
                
 \hline
\end{tabular}
\caption{Simulated annealing parameters for Lorenz '63 DR hyperparameter tuning}
\label{table:SimAnnealLorenz}
\end{table}

\begin{table}
\centering
\begin{tabular}{||l | l | l ||} 
 \hline
 DR	       &    Param Type                           &   Param \\ 
 \hline\hline
 PCA		       &                                         &   \\
 KPCA           &   kernel; gamma                         &  rbf; 0.95 \\
 UMAP           &   min\_dist; n\_neighbors	             &  0.02; 209 \\
 AE             &   batch\_size                           &  128 \\
 BNAE	       &   batch\_size                           &  128 \\
 CRAE	       &   batch\_size                           &  128 \\
 VAE	           &   batch\_size; KL\_target &  128; 2e-5\\
 DIRESA\textsubscript{MSE}        &   batch\_size; dist\_loss\_weight; &  512; 2.5; \\
                &   anneal\_step; cov\_target&  0.2; 2e-5\\
 DIRESA\textsubscript{Corr}       &   batch\_size; dist\_loss\_weight; &  512; 1.0; \\
                &   anneal\_step; cov\_target&  0.2; 2e-5\\
 All ANN's      &   en(de)coder hidden layers & dense [40, 20] \\
                &   activation; optimizer; & relu\slash linear; adam; \\
                &   lr\_start; epochs &  0.005; 200; \\
                &   lr\_decay & 0.5 every 10 epoch\\
                
 \hline
\end{tabular}
\caption{DR Configurations for the Lorenz '63 dataset}
\label{table:ModelConfLorenz}
\end{table}

\begin{table}
\centering
\begin{tabular}{||l | l | l | l ||} 
 \hline
 Layer       &    Output Shape          & Activation      &   Nbr of Params \\ 
 \hline\hline
 Input                 &   (3)             &                 &  0 \\
   \hline
 Enc\_Dense1           &   (40)            &   ReLu          &  160 \\
 Enc\_Dense2           &   (20)            &                 &  820 \\
 Enc\_Out\_Dense3      &   (2)             &   Linear        &  42 \\ 
  \hline
 Dec\_Dense3	          &   (20)            &   ReLu          &  60 \\
 Dec\_Dense2	          &   (40)            &   ReLu          &  840 \\
 Dec\_Out\_Dense1      &   (3)             &   Linear        &  123 \\
                
 \hline
\end{tabular}
\caption{ANN layers for the Lorenz '63 dataset}
\label{table:ModelDiresaLorenz}
\end{table}

For KernelPCA and UMAP, the hyperparameter tuning is done using simulated annealing. The hyperparameter tuning uses simulated annealing, which is an optimization metaheuristic. The name refers to the annealing process in metallurgy, where controlled cooling is done to get the material's right physical properties. The lower the temperature, the lesser the movements of the atoms. Simulated annealing is a local search heuristic, looking for a better solution in the neighborhood a the previous solution, which jumps from time to time to less good solutions to get out of local minima. The chance of such a jump lowers when the temperature is decreasing. The python \textit{simanneal} package is used, which requires the initial temperature T\textsubscript{max}, the final temperature T\textsubscript{min}, and the number of iterations to be set next to the initial values and the step size (see Table \ref{table:SimAnnealLorenz}. In our implementation, the step size is taken as a sample out of a normal distribution with a mean of 0 and a step standard deviation. The initial values were chosen based on the limited number of manual tuning runs to keep the number of iterations relatively low. The step standard deviation has been fixed to one-fifth of the initial values in order to have sufficient small steps and, at the same time, a sufficient reach for the parameter. The number of iterations is 200 for UMAP and  100 for KPCA (because of the high compute time). The annealing temperatures have been chosen in order to jump 10 percent of the time to less good solutions at the initial iterations and to almost zero at the end.  

The hyperparameters for the nine different DR techniques are listed in Table \ref{table:ModelConfLorenz}. KernelPCA has been tuned using a Gaussian kernel to an optimal gamma of 0.95. For UMAP, the values for min\_dist and nbr\_neighbors are 0.02 and 209. The ANN methods have an encoder submodel with a 3-dimensional input layer and two hidden layers with 40 and 20 fully connected nodes, see Table \ref{table:ModelDiresaLorenz}. The output layer has two fully connected nodes corresponding to the size of the latent space. The decoder submodel mirrors the encoder. In the BNAE, the 2-dimensional fully connected layer in the encoder is followed by a batch normalization layer. In the VAE, the encoder has 3 (2-dimensional) output layers (for mean, variance, and sample). The hidden layers use a \textit{relu} activation function, while the output layers use a \textit{linear} one. 

All methods are trained ten times (PCA and KernelPCA only once) with the training dataset, and the one with the smallest overall loss on the validation dataset has been chosen. The batch size for the ANNs is 128, except for DIRESA and CRAE, where it is 512, to have a covariance loss (and correlation distance loss) per batch, which is in the same order as the loss calculated over the total dataset. 
For the ANNs with multiple loss functions, the weight factor for the reconstruction loss is fixed to 1. For the VAE's KL loss weight factor, KL annealing is used, meaning that the weight factor is 0 at the start and gradually increases. The same annealing approach is used for the covariance loss weight factor for DIRESA and CRAE. The step to increase the annealing has been tuned to 0.2. Annealing stops when the covariance loss (or KL loss) reaches a target of 0.00002. At this point, the covariance of the latent components is below 0.005, and we consider them independent. Adam is used as the optimizer for the ANNs, and training is done during 200 epochs. The learning rate is divided by two every $10^{th}$ epoch after annealing is stopped or after the $50^{th}$ epoch for the ANN method without annealing.

\subsection*{MAOOAM DIRESA hyperparameters}

\begin{table}
\centering
\begin{tabular}{||l | l | l | l ||} 
 \hline
 Layer       &    Output Shape          & Activation      &   Nbr of Params \\ 
 \hline\hline
 Input                 &   (6, 8, 4)       &                 &  0 \\
 Shuffled\_Input       &   (6, 8, 4)       &                 &  0 \\
  \hline
 Enc\_Conv2D(3,3)      &   (6, 8, 100)     &   ReLu          &  3700 \\
 Enc\_MaxPooling2D     &   (3, 4, 100))    &                 &  0 \\
 Enc\_Flatten	      &   (1200)          &                 &  0 \\
 Enc\_Dense1	          &   (300)           &   ReLu          &  360300 \\
 Enc\_Dense2	          &   (100)           &   ReLu          &  30100 \\
 Enc\_Out\_Dense3      &   (10)            &   Linear        &  1010 \\ 
  \hline
 Dist\_Out             &   (2)             &                 &  0 \\ 
  \hline
 Dec\_Dense3	          &   (100)           &   ReLu          &  1100 \\
 Dec\_Dense2	          &   (300)           &   ReLu          &  30300 \\
 Dec\_Dense1	          &   (1200)          &   ReLu          &  361200 \\
 Dec\_Reshape          &   (3, 4, 100)     &                 &  0 \\
 Dec\_UpSampling2D     &   (6, 8, 100)     &                 &  0 \\
 Dec\_Out\_Conv2D(3,3) &   (6, 8, 4)       &   Linear        &  3604 \\
                
 \hline
\end{tabular}
\caption{DIRESA layers for the MAOOAM dataset}
\label{table:ModelDiresaMA}
\end{table}

\begin{table}
\centering
\begin{tabular}{|| l | l ||} 
 \hline
 Parameter                           &   Value \\ 
 \hline\hline
 
    batch\_size        &  512  \\
    dist\_loss\_weight &  1.0  \\
    anneal\_step       &  0.2  \\
    cov\_target        &  6.5e-5 \\
    optimizer          &  adam \\
    epochs             &  200 \\
    lr\_start          &  0.005  \\
    lr\_decay          &  0.5 every 10 epoch\\
                
 \hline
\end{tabular}
\caption{DIRESA hyperparameters for the MAOOAM dataset}
\label{table:ModelConfMA}
\end{table}

The DIRESA layers are listed in Table \ref{table:ModelDiresaMA}. The input layers have a shape of (8, 6, 4), corresponding to the four variables in each grid point. The encoder has one convolutional layer (with a 3 by 3 kernel), followed by a MaxPooling (with a two by two pooling size), a Flatten, and three fully connected layers. The last one has ten nodes corresponding to the size of the latent space. The decoder submodel mirrors the encoder. The hidden layers use a \textit{relu} activation function, while the encoder and decoder output layers use a \textit{linear} one. The third output has two values: the distance between the two inputs and between the two latent representations. The weight factors for the reconstruction and distance losses are fixed to 1, and for the covariance loss, weight annealing is used. 

The DIRESA methods are trained ten times with the training dataset, and the one with the smallest overall loss on the validation dataset has been chosen. The batch size is 512; see Table  \ref{table:ModelConfMA}. The weight factors for the reconstruction and distance losses are fixed to 1, and for the covariance loss, weight annealing is used. The step to increase the annealing is set to 0.2. Annealing stops when the covariance loss reaches a target of 0.000065. At this point, we consider the latent components independent. Adam is used as the optimizer, and training is done during 200 epochs. The learning rate is divided by two every $10^{th}$ epoch after annealing is stopped.

\section{Distance ordering preservation}\label{sec:dist}
\subsection*{Lorenz '63 distance preservation}

\begin{table}
\centering
\begin{tabular}{|l || l | l || l | l | l | l |} 
\hline
Model & Corr & LogCorr & Can50 & Pear50 & Spear50 & Ken50 \\
\hline\hline
PCA & 5.10e-5 & 5.86e-5 & 0.00169 & 0.00305 & 0.00215 & 0.00210 \\
\hline
KernelPCA & 0.000112 & 0.000108 & 0.00156 & 0.00284 & 0.00203 & 0.00189 \\
\hline
UMAP & 0.000503 & 0.000409 & 0.00127 & 0.00189 & 0.00170 & 0.00158 \\
\hline

AE & 0.000529 & 0.000439 & 0.00109 & 0.00160 & 0.00141 & 0.00134 \\
\hline
BNAE & 0.000141 & 0.000119 & 0.00146 & 0.00249 & 0.00192 & 0.00188 \\
\hline
CRAE & 0.000493 & 0.000469 & 0.00125 & 0.00195 & 0.00148 & 0.00156 \\
\hline

VAE & 0.000118 & 0.000150 & 0.00142 & 0.00252 & 0.00191 & 0.00187 \\
\hline
DIRESA\textsubscript{MSE} & 4.80e-5 & 5.32e-5 & 0.00167 & 0.00298 & 0.00211 & 0.00210 \\
\hline
DIRESA\textsubscript{Corr} & 4.66e-5 & 5.30e-5 & 0.00168 & 0.00302 & 0.00212 & 0.00210 \\
\hline

\hline
\end{tabular}
\caption{Distance preservation standard error of mean for the Lorenz '63 test dataset. Columns as in Table \ref{table:LorenzDistPreservationMean50}.}
\label{table:LorenzDistPreservationStdErr50}
\end{table}

\begin{table}
\centering
\begin{tabular}{|l || l | l | l | l |} 
\hline
Models & Can50 & Pear50 & Spear50 & Ken50 \\
\hline\hline

DIRESA\textsubscript{MSE} - PCA &  2.59e-05 & 0.019 & 0.020 & 2.46e-06 \\
\hline

DIRESA\textsubscript{Corr} - PCA & 0.036 & \textbf{0.162} & \textbf{0.098} & 0.002 \\
\hline

\end{tabular}
\caption{2 sample t-test p-values for the mean distance between DIRESA and PCA for the Lorenz test dataset. Columns as in Table \ref{table:LorenzDistPreservationMean50}. Values below 5\% are marked in bold.}
\label{table:LorenzDistPreservationPvalue}
\end{table}

\begin{figure}[ht]
\centering
\includegraphics[width=\textwidth]{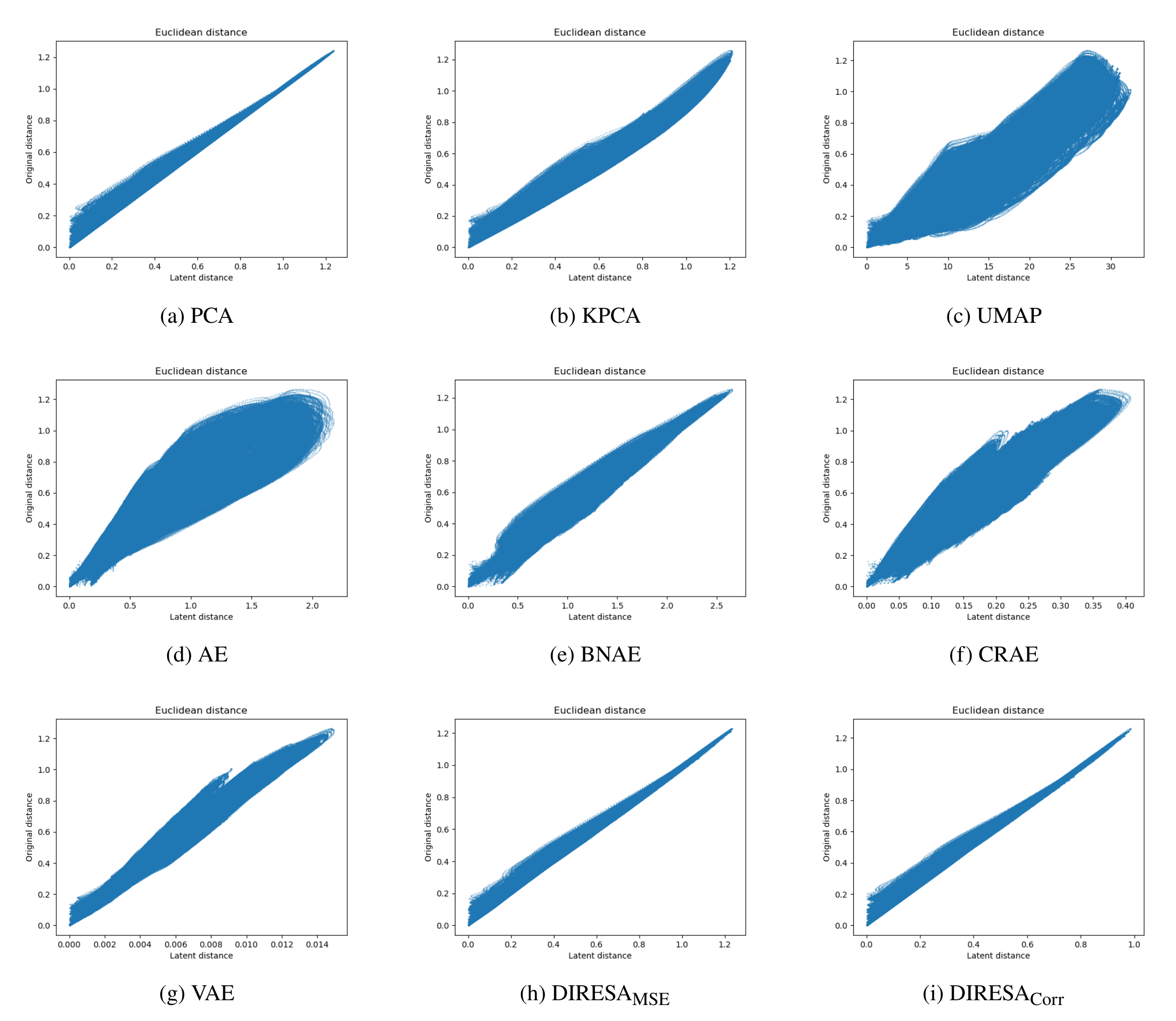}
\caption{Scatter plots of distances for 200 random samples of the Lorenz test dataset for the different DR techniques} 
\label{fig:LorenzCorr}
\end{figure}

\begin{table}
\centering
\begin{tabular}{|l || l | l | l | l |} 
\hline
DR &  Can100 & Pear100 & Spear100 & Ken100 \\
\hline\hline

PCA &  \textbf{0.115} & 0.854 & \textbf{0.894} & \textbf{0.836} \\
\hline
KernelPCA &  0.199 & 0.791 & 0.826 & 0.706 \\
\hline
UMAP &  0.271 & 0.749 & 0.774 & 0.624 \\
\hline

AE &  0.249 & 0.821 & 0.811 & 0.654 \\
\hline
BNAE &  0.148 & \textbf{0.874\textsuperscript{*}} & 0.883 & 0.773 \\
\hline
CRAE &  0.174 & \textbf{0.874} & 0.875 & 0.745 \\
\hline

VAE &  0.149 & 0.868 & 0.878 & 0.768 \\
\hline
DIRESA\textsubscript{MSE} & \textbf{0.105\textsuperscript{*}} & \textbf{0.864} & \textbf{0.901\textsuperscript{*}} & \textbf{0.850\textsuperscript{*}} \\
\hline
DDIRESA\textsubscript{Corr} & \textbf{0.109} & 0.859 & \textbf{0.899} & \textbf{0.846} \\
\hline

\end{tabular}
\caption{Mean distance ordering preservation between original and latent space for the Lorenz '63 test dataset. The \textit{Can100} column gives the Canberra stability indication (smaller is better), the \textit{Pear100}, \textit{Spear100} and \textit{Ken100} show the Pearson, Spearman, and Kendall correlation with a location parameter of 100. The 3 best values are highlighted in bold;  the best one is marked with an asterisk.}
\label{table:LorenzDistPreservationMean100}
\end{table}

\begin{table}
\centering
\begin{tabular}{|l || l | l | l | l |} 
\hline
DR &  Can100 & Pear100 & Spear100 & Ken100 \\
\hline\hline

PCA  &  \textbf{0.050} & \textbf{0.991} & \textbf{0.988} & \textbf{0.920} \\
\hline
KernelPCA  &  0.158 & 0.894 & 0.893 & 0.746 \\
\hline
UMAP  &  0.253 & 0.771 & 0.798 & 0.629\\
\hline

AE  &  0.229 & 0.846 & 0.842 & 0.673 \\
\hline
BNAE  &  0.097 & 0.971 & 0.955 & 0.837 \\
\hline
CRAE  &  0.155 & 0.930 & 0.912 & 0.763 \\
\hline

VAE  &  0.105 & 0.966 & 0.951 & 0.827 \\
\hline
DIRESA\textsubscript{MSE}  & \textbf{0.040\textsuperscript{*}} & \textbf{0.995\textsuperscript{*}} & \textbf{0.992\textsuperscript{*}} & \textbf{0.935\textsuperscript{*}} \\
\hline
DIRESA\textsubscript{Corr}  & \textbf{0.044} & \textbf{0.994} & \textbf{0.990} & \textbf{0.929} \\
\hline

\end{tabular}
\caption{Median distance ordering preservation between original and latent space for the Lorenz '63 test dataset. Columns and highlights as in Table \ref{table:LorenzDistPreservationMean100}.}
\label{table:LorenzDistPreservationMedian100}
\end{table}

\begin{table}
\centering
\begin{tabular}{|l || l | l | l | l |} 
\hline
Model &  Can100 & Pear100 & Spear100 & Ken100 \\
\hline\hline
PCA &  0.00165 & 0.00281 & 0.00208 & 0.00205 \\
\hline
KernelPCA &  0.00148 & 0.00259 & 0.00193 & 0.00178 \\
\hline
UMAP &  0.00119 & 0.00173 & 0.00157 & 0.00147 \\
\hline

AE &  0.00097 & 0.00125 & 0.00128 & 0.00121 \\
\hline
BNAE &  0.00133 & 0.00210 & 0.00179 & 0.00177 \\
\hline
CRAE &  0.00115 & 0.00162 & 0.00132 & 0.00144 \\
\hline

VAE & 0.00136 & 0.00229 & 0.00185 & 0.00182 \\
\hline
DIRESA\textsubscript{MSE} &  0.00163 & 0.00273 & 0.00204 & 0.00204 \\
\hline
DIRESA\textsubscript{Corr} &  0.00164 & 0.00277 & 0.00205 & 0.00205 \\
\hline

\hline
\end{tabular}
\caption{Distance preservation standard error of mean for the Lorenz '63 test dataset. Columns as in Table \ref{table:LorenzDistPreservationMean100}.}
\label{table:LorenzDistPreservationStdErr100}
\end{table}

\begin{figure}[ht]
\centering
\includegraphics[width=\textwidth]{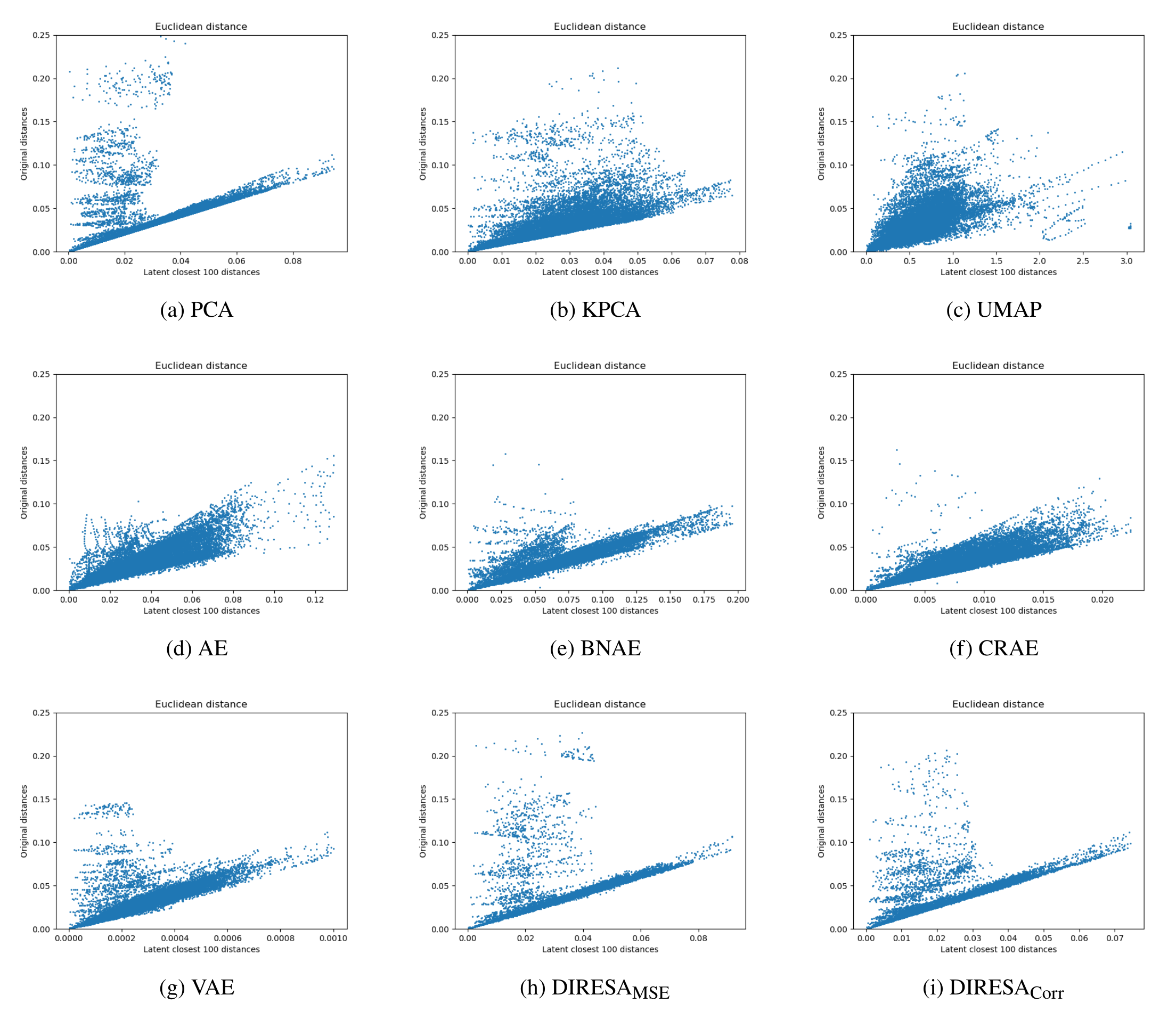}
\caption{Scatter plots of 100 closest distances for 200 random samples of the Lorenz test dataset for the different DR techniques} 
\label{fig:LorenzCorr100}
\end{figure}

Table \ref{table:LorenzDistPreservationStdErr50} shows the standard error of the mean for the \textit{Corr} and \textit{LogCorr} distance KPIs and for the KPIs with a location parameter of 50: \textit{Can50}, \textit{Pear50}, \textit{Spear50} and \textit{Ken50}. Table \ref{table:LorenzDistPreservationPvalue} shows the 2 sample t-test p-values for the mean distance between DIRESA and PCA. Figure \ref{fig:LorenzCorr} shows the scatter plots of the distances (latent versus original space) for the different methods for 200 random samples of the test dataset.

Tables \ref{table:LorenzDistPreservationMean100} and \ref{table:LorenzDistPreservationMedian100} show how distance ordering is preserved between original and latent space when a location parameter of 100 is used. For the mean scores, DIRESA\textsubscript{MSE} scores best on 3 out of 4 indicators, while for the median, it scores best on all 4. Tables \ref{table:LorenzDistPreservationStdErr100} show the standard error of the mean for the KPIs with a location parameter of 100. Figure \ref{fig:LorenzCorr100} shows the scatter plots of the 100 closest distances (in latent space) for the different methods for 200 random samples of the test dataset.

\subsection*{MAOOAM distance preservation}

\begin{table}
\centering
\begin{tabular}{|l || l | l || l | l | l | l |} 
\hline
Model & Corr & LogCorr & Can50 & Pear50 & Spear50 & Ken50 \\
\hline\hline
PCA & 0.000 & 0.000 & 0.009 & 0.013 & 0.013 & 0.010 \\
\hline

DIRESA\textsubscript{MSE} & 0.000 & 0.000 & 0.003 & 0.014 & 0.014 & 0.011 \\
\hline
DIRESA\textsubscript{MSLE} & 0.000 & 0.000 & 0.011 & 0.015 & 0.015 & 0.013 \\
\hline
DIRESA\textsubscript{Corr} & 0.000 & 0.000 & 0.009 & 0.014 & 0.014 & 0.011 \\
\hline
DIRESA\textsubscript{LogCorr} & 0.000 & 0.000 & 0.009 & 0.013 & 0.014 & 0.012 \\
\hline

\hline
\end{tabular}
\caption{Distance preservation standard error of mean for 200 random samples of the MAOOAM test dataset. Columns as in Table \ref{table:LorenzDistPreservationMean50}.}
\label{table:MADistPreservationStdErr}
\end{table}

\begin{table}
\centering
\begin{tabular}{|l || l | l | l | l |} 
\hline
Models & Can50 & Pear50 & Spear50 & Ken50 \\
\hline\hline

DIRESA\textsubscript{MSLE} - PCA &  0.0003 & 0.004 & 0.004 & 0.006 \\
\hline

DIRESA\textsubscript{LogCorr} - PCA & 0.005 & 0.031 & 0.024 & 0.022 \\
\hline

\end{tabular}
\caption{2 sample t-test p-values for the mean distance between DIRESA and PCA for the MAOOAM test dataset. Columns as in Table \ref{table:LorenzDistPreservationMean50}. }
\label{table:MADistPreservationPvalue}
\end{table}

Table \ref{table:MADistPreservationStdErr} shows the standard error of the mean for the \textit{Corr} and \textit{LogCorr} distance KPIs and for the KPIs with a location parameter of 50: \textit{Can50}, \textit{Pear50}, \textit{Spear50} and \textit{Ken50}. Table \ref{table:MADistPreservationPvalue} shows the 2 sample t-test p-value for the mean distance between DIRESA and PCA.

\section{Ordering and visualisation of latent variables} \label{sec:hierarchy}

\subsection*{Latent variable ordering procedure}

The ordering of the encoded or latent variables is done by ranking them according to the $R^2$ score they contribute when they are decoded back into the original space. The following procedure was applied to assess this contribution for each variable. 

We start by encoding the original dataset $\mathbf{X}= \left(\mathbf{x}_i\right)_{i=1\ldots n}$ with the encoder $f_{\theta}$ to yield the encoded (latent) dataset $ f_{\theta}(\mathbf{X}) = \mathbf{Z} \in \mathbb{R}^{n \times L}$, where $n$ is the number of data points, and $L$ is the number of latent components. The latent representation of the $i$-th data point is $ \mathbf{z}_i = (z_{i1}, z_{i2}, \dots, z_{iL})$. 
For each component \(l\), we generate a new dataset $\mathbf{Z}^{(l)}$ by preserving the values of the \(l\)-th component and setting the values of all other components $j$ to their respective mean values $\mu_j$:
\begin{align} \mathbf{Z}^{(l)} = \left( z_{ij}^{(l)} \right)_{i=1,\dots,n, \, j=1,\dots,L} \quad \text{where} \quad
z_{ij}^{(l)} = 
\begin{cases} 
z_{il} & \text{if } j = l \\
\mu_j & \text{if } j \neq l
\end{cases}
\end{align}
yielding \(L\) new datasets: \( \mathbf{Z}^{(1)}, \mathbf{Z}^{(2)}, \dots, \mathbf{Z}^{(L)} \).

In a second step, each new dataset is decoded back into the original space with the decoder function $g_\phi$             , yielding a corresponding decoded dataset \( \mathbf{\hat{X}}^{(l)} \):
\[
\mathbf{\hat{X}}^{(l)} = g_{\phi}(\mathbf{Z}^{(l)}).
\]

In the third step, the $R^2_{(l)}$ score in the original space is computed for each component
\[
R^2_{(l)} = 1 - \frac{\sum_{i=1}^{n} \left\| \mathbf{x}_i -  \mathbf{\hat{x}}_i^{(l)}  \right\|^2}{\sum_{i=1}^{n} \left\| \mathbf{x}_i -  \mathbf{\bar{x}}  \right\|^2} 
\]
where $\mathbf{\hat{x}}_i^{(l)}$ is the $i$ -th data point in the decoded dataset \( \mathbf{\hat{X}}^{(l)} \).

Finally, the components are ranked in decreasing order of their corresponding $R^2_{(l)}$ scores, which yields an ordered list of components, where the first component contributes the most, and the last component contributes the least. This procedure generalizes the ordering according to the explained variance of principal components to nonlinear dimension reduction techniques. 

\subsection*{Cumulative component $R^2_{(l)}$ scores}

Due to the nonlinearity, we can not sum the $R^2_{(l)}$ scores of the ranked components. For calculating the cumulative $R^2_{(l)}$ scores, we start from the ordered latent representation of the $i$-th data point $\mathbf{z}_{ij}$ ($j$ being the index of the ordered component). The procedure is equal to the one explained in the previous paragraph, except that Equation D1 is replaced with
\begin{align} \mathbf{Z}^{(l)} = \left( z_{ij}^{(l)} \right)_{i=1,\dots,n, \, j=1,\dots,L} \quad \text{where} \quad
z_{ij}^{(l)} = 
\begin{cases} 
z_{il} & \text{if } j \leq l \\
\mu_j & \text{if } j > l
\end{cases}
\end{align}
for $l < L$. In case $l = L$, the cumulative $R^2_{(l)}$ is equal to $R^2$.

\subsection*{Latent variable interpretation}

Similar to the ordering procedure, we can evaluate the meaning of a latent variable by measuring the effect of varying the variable in question while setting the others at their mean value. For each latent component $l$, we construct two latent vectors, in which the $l$-th component is set to the mean plus or minus one standard deviation
\[
\mathbf{z}^{(l,\pm)} = \left(z_{j}^{(l,\pm)}\right)_{j=1,\dots,L} \quad \text{where} \quad
z_{j}^{(l,\pm)} = 
\begin{cases} 
\mu_l \pm \sigma_l & \text{if } j = l \\
\mu_j & \text{if } j \neq l.
\end{cases}
\]

Decoding these vectors into the original space using $g_\phi$ allows us to visualize the effect of $l$:

\[
\hat{\mathbf{x}}^{(l,+)} = g_{\phi}(\mathbf{z}^{(l,+)}), \quad \hat{\mathbf{x}}^{(l,-)} = g_{\phi}(\mathbf{z}^{(l,-)}).
\]

The difference $ \Delta \hat{\mathbf{x}}^{(l)} = \hat{\mathbf{x}}^{(l,+)} - \hat{\mathbf{x}}^{(l,-)}$ approximately captures the influence of the latent variable $l$. As a caveat, this method is approximative and may be inaccurate in the presence of strong nonlinear interactions between latent variables, which are minimized for DIRESA thanks to the covariance regularization. 

\section{Ablation analysis}\label{sec:ablation}

\begin{figure}[h]
\centering
\includegraphics[width=0.8\textwidth]{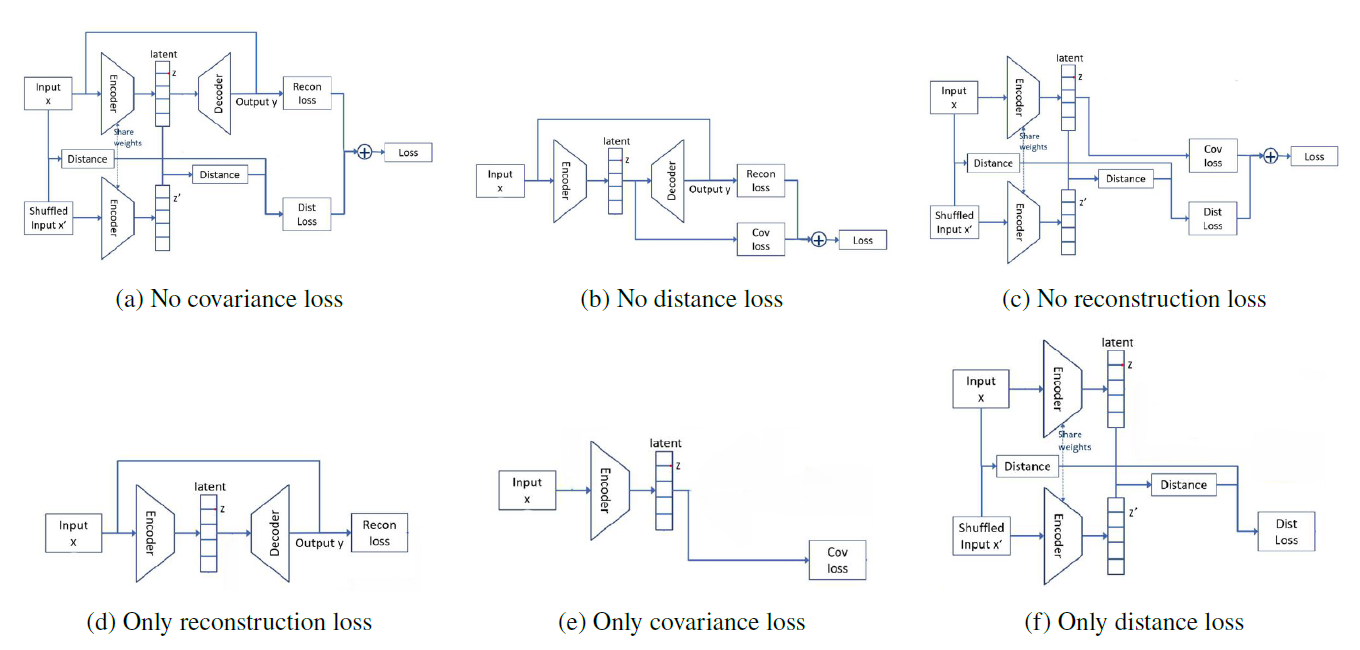}
\caption{Different ablation scenarios}
\label{fig:ablation}
\end{figure}

In this section, we will quantitatively demonstrate each loss function's usefulness. Six different ablation scenarios have been investigated (see Figure \ref{fig:ablation}):
    \begin{enumerate}[(a)]
        \item No covariance loss
        \item No distance loss, this is the CRAE as investigated in the paper
        \item No reconstruction loss
        \item Only reconstruction loss, this is the AE as investigated in the paper
        \item Only covariance loss
        \item Only reconstructions loss
    \end{enumerate}

\subsection*{Lorenz '63 distance preservation}

\begin{figure}[ht]
\centering
\includegraphics[width=\textwidth]{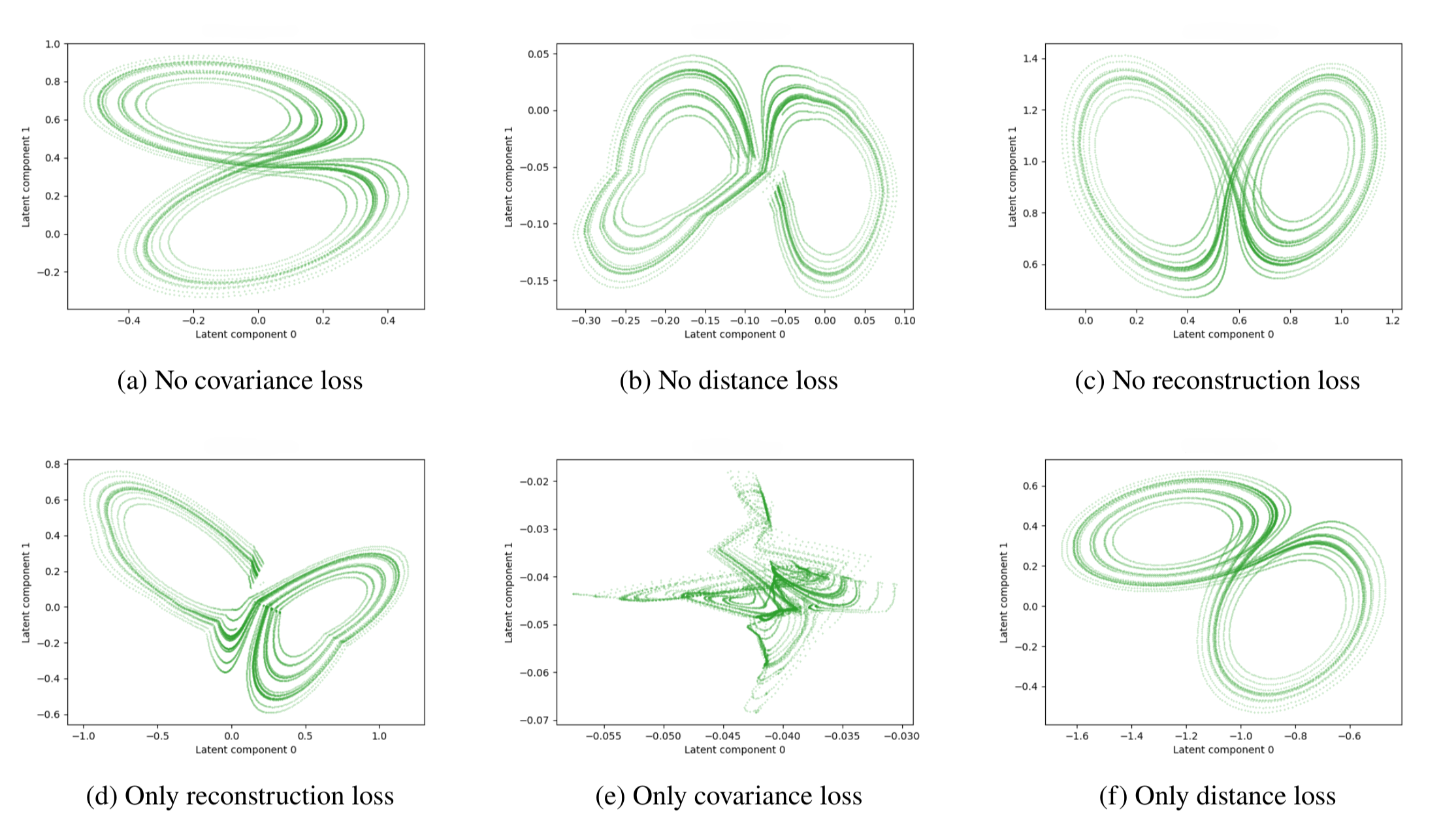}
\caption{Encoded Lorenz test dataset for different ablation scenarios}
\label{fig:latentablation}
\end{figure}

Figure \ref{fig:latentablation} shows the mapping of the Lorenz test dataset into latent space for the different ablation scenarios. An MSE distance loss function is used, and the hyperparameters are set as in Table \ref{table:ModelConfLorenz}. For the scenarios without a covariance loss (Figure \ref{fig:latentablation} a, d, and f), the encoded latent components are correlated.  The wings are distorted for the scenarios without a distance loss (Figure \ref{fig:latentablation} b, d and e). For the scenario with a covariance and distance loss but without a reconstruction loss (Figure \ref{fig:latentablation} c), the latent mapping is equal to the one with a reconstruction loss (see Figure \ref{fig:latentlorenz} h and i). This shows that the reconstruction loss is unnecessary if only the latent encoding is required. Even the scenario with only a distance loss can be considered if there is no need for having independent latent components (Figure \ref{fig:latentablation} f). The wings are completely distorted in the scenario with only a covariance loss (Figure \ref{fig:latentablation} e).

Table \ref{table:LorenzDistPreservationAblationMean50} and \ref{table:LorenzDistPreservationAblationMedian50} show the distance (ordering) preservation KPIs (mean and median for 200 random samples) for the different ablation scenarios and the DIRESA method with all three loss functions. In the scenario with only a covariance loss, all the KPIs are the worst, making it useless. The scenarios without a distance loss (b, d) are doing worse than those with a distance loss. This shows the usefulness of distance loss in enforcing distance (ordering) preservation. The scenarios with distance loss and without reconstruction loss (c and f) have comparable results to the full DIRESA method, showing that the reconstruction loss is unnecessary for distance (ordering) preservation. 

\begin{table}
\centering
\begin{tabular}{|l || l | l || l | l | l | l |} 
\hline
DR & Corr & LogCorr & Can50 & Pear50 & Spear50 & Ken50 \\
\hline\hline

a. No covariance loss & 0.996 & 0.996 & 0.128 & 0.816 & 0.877 & 0.824 \\
\hline
b. No distance loss & 0.945 & 0.946 & 0.178 & \textbf{0.859\textsuperscript{*}} & 0.866 & 0.741 \\
\hline
c. No reconstruction loss & \textbf{0.997\textsuperscript{*}} & \textbf{0.997\textsuperscript{*}} & \textbf{0.117} & \textbf{0.856} & \textbf{0.897} & \textbf{0.841} \\
\hline

d. Only reconstruction loss & 0.924 & 0.936 & 0.257 & 0.805 & 0.800 & 0.649 \\
\hline
e. Only covariance loss & 0.353 & 0.415 & 0.613 & 0.207 & 0.352 & 0.265 \\
\hline
f. Only distance loss & \textbf{0.997\textsuperscript{*}} & \textbf{0.997\textsuperscript{*}} & \textbf{0.105\textsuperscript{*}} & \textbf{0.851} & \textbf{0.907\textsuperscript{*}} & \textbf{0.854\textsuperscript{*}} \\
\hline

Full DIRESA & \textbf{0.997\textsuperscript{*}} & \textbf{0.997\textsuperscript{*}} & \textbf{0.107} & 0.849 & \textbf{0.896} & \textbf{0.848} \\
\hline

\end{tabular}
\caption{Mean distance ordering preservation between original and latent space for 200 random samples for the Lorenz '63 test dataset for the different ablation scenarios and the DIRESA method withh all three loss functions. Columns and highlights as in Table \ref{table:LorenzDistPreservationMean50}.}
\label{table:LorenzDistPreservationAblationMean50}
\end{table}

\begin{table}
\centering
\begin{tabular}{|l || l | l || l | l | l | l |} 
\hline
DR & Corr & LogCorr & Can50 & Pear50 & Spear50 & Ken50 \\
\hline\hline

a. No covariance loss & \textbf{0.999\textsuperscript{*}} & \textbf{0.999\textsuperscript{*}} & 0.052 & 0.993 & 0.986 & 0.918 \\
\hline
b. No distance loss & 0.964 & 0.964 & 0.151 & 0.930 & 0.914 & 0.768 \\
\hline
c. No reconstruction loss & \textbf{0.999\textsuperscript{*}} & \textbf{0.999\textsuperscript{*}} & \textbf{0.050} & \textbf{0.994} & \textbf{0.990\textsuperscript{*}} & \textbf{0.931} \\
\hline

d. Only reconstruction loss & 0.938 & 0.947 & 0.233 & 0.845 & 0.833 & 0.670 \\
\hline
e. Only covariance loss & 0.410 & 0.453 & 0.629 & 0.169 & 0.367 & 0.256 \\
\hline
f. Only distance loss & \textbf{0.999\textsuperscript{*}} & \textbf{0.999\textsuperscript{*}} & \textbf{0.046} & \textbf{0.994} & \textbf{0.990\textsuperscript{*}} & \textbf{0.931} \\
\hline

Full DIRESA & \textbf{0.999\textsuperscript{*}} & \textbf{0.999\textsuperscript{*}} & \textbf{0.040\textsuperscript{*}} & \textbf{0.995\textsuperscript{*}} & \textbf{0.990\textsuperscript{*}} & \textbf{0.935\textsuperscript{*}} \\
\hline

\end{tabular}
\caption{Median distance ordering preservation between original and latent space for 200 random samples for the Lorenz '63 test dataset for the different ablation scenarios and the DIRESA method withh all three loss functions. Columns and highlights as in Table \ref{table:LorenzDistPreservationMean50}.}
\label{table:LorenzDistPreservationAblationMedian50}
\end{table}

\subsection*{MAOOAM distance preservation}

As for the Lorenz dataset, the ablation scenarios with distance loss and without reconstruction loss show the best distance (ordering) preservation KPIs; we will only use these for the MAOOAM dataset. We will, however, now investigate a second distance loss function, the MSLE (as this one gave the best results in Table \ref{table:MADistPreservation}). The hyperparameters are kept the same as in Table \ref{table:ModelConfMA}.

Table \ref{table:MADistPreservationAblation} shows the distance (ordering) preservation KPIs (mean for 200 random samples) for the ablation scenarios and the DIRESA method with all three loss functions. The scenario with only an MSLE distance loss is the best for all the defined KPIs, followed by the full DIRESA with an MSLE distance loss. When having only a distance loss function, no trade-off has to be made between the different loss functions.

It is important to note that distance in this section is a distance over the four MAOOAM variables, atmospheric and ocean streamfunction and temperature. 

However, when searching for analogs for a particular variable or group of variables (e.g., MAOOAM ocean analogs), a decoder is needed to detect the latent components that impact that variable or group of variables. The MAOOAM ocean streamfunction cumulative $R^2_{(l)}$ for the first three latent components is 98\%. So, when looking for ocean streamfunction analogs, a distant search can be performed on only the first three components of the latent dataset. When looking for the ocean analogs, streamfunction, and temperature, the fourth component needs also be taken into account (as this one has an $R^2_{(l)}$ of 25\% for the ocean temperature).

\begin{table}
\centering
\begin{tabular}{|l || l | l || l | l | l | l |} 
\hline
DR & Corr & LogCorr & Can50 & Pear50 & Spear50 & Ken50 \\
\hline\hline

No reconstruction loss\textsubscript{MSE} & \textbf{0.999\textsuperscript{*}} & \textbf{0.998} & 0.569 & 0.418 & 0.390 & 0.282 \\
\hline
No reconstruction loss\textsubscript{MSLE} & 0.998 & 0.997 & 0.599 & 0.396 & 0.369 & 0.268 \\
\hline

Only distance loss\textsubscript{MSE} & \textbf{0.999\textsuperscript{*}} & \textbf{0.998} & \textbf{0.562} & \textbf{0.441} & \textbf{0.401} & 0.290 \\
\hline
Only distance loss\textsubscript{MSLE} & \textbf{0.999\textsuperscript{*}} & \textbf{0.999\textsuperscript{*}} & \textbf{0.503\textsuperscript{*}} & \textbf{0.501\textsuperscript{*}} & \textbf{0.464\textsuperscript{*}} & \textbf{0.345\textsuperscript{*}} \\
\hline

DIRESA\textsubscript{MSE} & \textbf{0.999\textsuperscript{*}} & \textbf{0.999\textsuperscript{*}} & 0.565 & 0.429 & \textbf{0.401} & \textbf{0.292} \\
\hline
DIRESA\textsubscript{MSLE} & \textbf{0.999\textsuperscript{*}} & \textbf{0.998} & \textbf{0.518} & \textbf{0.490} & \textbf{0.453} & \textbf{0.338} \\       
\hline

\end{tabular}
\caption{Mean distance ordering preservation between original and latent space for 200 random samples of the MAOOAM test dataset for different ablation scenarios. Columns and highlights as in Table \ref{table:LorenzDistPreservationMean50}.}
\label{table:MADistPreservationAblation}
\end{table}

\clearpage 

\bibliography{References}

\end{document}